\documentclass{article}

\usepackage{open_preprint}
\renewcommand{\preprintnotice}{}

\usepackage[utf8]{inputenc}
\usepackage[T1]{fontenc}

\usepackage{
    amssymb,
    amsmath,
    amsfonts,
    eurosym,
    graphicx,
    comment,
    pdflscape,
    subfigure,
    array,
    longtable,
    dsfont,
    etoc,
    xcolor
}

\usepackage[normalem]{ulem}
\usepackage{bbm}
\usepackage{algorithm}
\usepackage{algpseudocode}
\usepackage{booktabs}
\usepackage{microtype}
\usepackage{url}
\usepackage{wrapfig}
\usepackage{titlesec}

\usepackage{titletoc}

\definecolor{RoyalBlue}{HTML}{5A678A}
\definecolor{MoBlue}{HTML}{18273E}
\definecolor{refblue}{HTML}{3B6DE2}
\definecolor{eqmagenta}{HTML}{B33B6A}
\definecolor{OptOrange}{HTML}{A94F1F}

\let\oldtextbf\textbf
\renewcommand{\textbf}[1]{%
    {\color{MoBlue}\oldtextbf{#1}}%
}

\titleformat{\paragraph}
    [runin]
    {\bfseries\color{MoBlue}}
    {}
    {0pt}
    {}

\usepackage{hyperref}

\hypersetup{
    pdftitle={OptScientist: Automated Discovery of Optimization Laws},
    pdfauthor={Ma ZhenYao, Yue Liang, Jian-Qiao Zhu},
    pdfdisplaydoctitle=true
}

\hypersetup{
    colorlinks=true,
    linkcolor=eqmagenta,
    citecolor=refblue,
    urlcolor=refblue
}

\newtheorem{theorem}{Theorem}[section]

\newtheorem{question}{Question}

\usepackage{authblk}

\newcolumntype{L}[1]
{>{\raggedright\let\newline\\\arraybackslash\hspace{0pt}}m{#1}}

\newcolumntype{C}[1]
{>{\centering\let\newline\\\arraybackslash\hspace{0pt}}m{#1}}

\newcolumntype{R}[1]
{>{\raggedleft\let\newline\\\arraybackslash\hspace{0pt}}m{#1}}

\title{ {\huge{\color{OptOrange}\bfseries Opt}%
    {\color{black}\bfseries Scientist}:}\\[0.5em] Automated Discovery of Optimization Laws}

\author[1,2]{{Ma ZhenYao}%
\thanks{Corresponding author: \href{mailto:zhenyaoma@163.com}{zhenyaoma@163.com}}%
\hspace{0.5em}%
}
\author[1,3]{Yue Liang}
\author[4]{Jian-Qiao Zhu}

\affil[1]{Plokerm AI Lab}
\affil[2]{Xiamen University}
\affil[3]{University of Tübingen}
\affil[4]{The University of Hong Kong}

\begin{document}

\vspace*{-0.8cm}
\maketitle

\begin{abstract}
Many scientific phenomena are best understood as outcomes of underlying optimization problems, yet automated scientific discovery has largely focused on recovering equations that describe observed input–output patterns. We introduce \textbf{OptScientist}, a novel automated framework for discovering compact symbolic optimization laws directly from data. OptScientist comprises (i) a model that infers symbolic optimization problems by first learning to represent data as an optimization problem and then extracting its symbolic form, and (ii) an LLM agentic system embedding the model in a discovery process that integrates theoretical priors from scientific literature, discovers candidate laws, abstracts their underlying mechanisms, and suggests new data to collect. We first evaluate OptScientist on \emph{OptFeynman}, a novel benchmark of 99 physics-derived optimization problems, where OptScientist recovers ground-truth laws for over 80\% of the problems and remains robust to noisy contextual information, while substantially outperforming machine-learning, symbolic, and LLM-based model-discovery baselines. On human risky choice and central-bank monetary policy, OptScientist discovers compact symbolic optimization laws that achieve held-out predictive performance competitive with leading baselines, while using few parameters and substantially fewer tokens than LLM baselines. Overall, OptScientist extends automated scientific discovery to uncovering optimization laws that generate observed phenomena, enabling automated theory building across the physical, biological, behavioral, and social sciences.
\end{abstract}

\section{Introduction} \label{sec:introduction}

Across the sciences, many influential theories characterize observed phenomena as solutions to optimization problems. Examples include energy or free-energy minimization principles for characterizing equilibrium states in statistical physics \citep{gibbs1902elementary, landau2013statistical}, optimality models for explaining how biological fitness shapes evolutionary outcomes \citep{wright1932roles, fisher1999genetical}, rational and resource-rational analyses of human behavior and cognition \citep{samuelson1948foundations, debreu1959theory, anderson1991adaptive, lieder2020resource}, and intertemporal optimization for modelling dynamic macroeconomic systems \citep{ramsey1928mathematical, ljungqvist2018recursive}. Developing such theories requires identifying objectives and relevant constraints that jointly account for observed outcomes. However, these components are typically formulated by domain experts and only subsequently evaluated against data, making optimization-based theory construction largely manual and dependent on specialized insight. This reliance on expert formulation creates a substantial bottleneck for theory discovery across scientific domains. A fundamental question therefore remains: \textit{Can optimization laws be discovered automatically from observational data?}

The complexity of real-world scientific data further exacerbates this bottleneck, as observations are often noisy, high-dimensional, and shaped by complex interactions. Hand-crafted interpretable models may fail to capture complex patterns, whereas flexible data-driven approaches, particularly machine-learning methods, have demonstrated strong predictive performance across physics \citep{baldi2014searching,carleo2017solving}, biology \citep{poplin2018universal,jumper2021highly}, and behavioral and social sciences \citep{kosinski2013private,peterson2021using,zhu2025capturing}. 

Recent advances in scientific machine learning have enabled data-driven discovery of symbolic laws \citep{champion2019data,udrescu2020ai,cranmer2020discovering}, embedded mathematical structure into flexible ML models \citep{chen2018neural,liu2025kan,ma2026behavior}, and integrated scientific theories with machine learning \citep{raissi2019physics,peterson2021using}. However, these approaches have largely focused on recovering input–output equations or learning models without explicit symbolic structure, while optimization-structured scientific ML models \citep{ma2026behavior} still require human interpretation of optimization structures and lack an end-to-end process for automated theory discovery. Consequently, automatically discovering compact symbolic optimization laws from real-world data remains a fundamental challenge.

To address these challenges, we introduce \textbf{OptScientist}, \textit{a novel framework that uses AI methods to automate the discovery of optimization laws.} OptScientist has two components: a novel symbolic-discovery model
and an agentic research system. First, the symbolic-discovery model (\emph{OptScientist Model}) learns to represent observational data as an optimization problem using a hierarchical optimization network with an optimization representation bottleneck, enabling denoising, information compression, and automatic extraction of interpretable features. It then distills this learned optimization representation into a compact symbolic optimization problem. The second component, the \emph{OptScientist Agent}, places this discovered symbolic model inside an
automated scientific-discovery loop. A Theory Prior Agent
proposes scientific priors, candidate variables and mechanism hypotheses. An Experiment Agent then uses the OptScientist Model to search for compact symbolic optimization laws from data. Finally, a Scientific Critic and Theory Abstraction Agent evaluates and selects the discovered laws, abstracts the selected laws into a general form and qualitative mechanisms, and suggests new variables or data for the next round of discovery. Together, the model and agent form an iterative system for proposing, discovering and refining optimization laws.

We conduct three experiments spanning physical systems, human behaviour, and macroeconomic policy. First, to test whether the core OptScientist Model can recover ground-truth optimization laws (Sec.~\ref{sec:expr-ground-truth}), we introduce \emph{OptFeynman} (Sec.~\ref{sec:optfeynman}), a novel synthetic benchmark of 99 optimization problems derived from analytic physics equations, where observations are generated as noisy samples from the induced optimization landscapes. On held-out contexts, OptScientist Model shows strong recovery of the generating laws, achieving perfect exact-order recovery on 88 of 99 discrete problems and 78 of 99 continuous problems, while substantially outperforming machine-learning, symbolic-regression, and LLM model-discovery baselines. Our results also suggest that OptScientist remains robust even when context variables are noisy. We then apply OptScientist to two autonomous scientific-discovery tasks on real-world data for which the underlying laws are unknown. In human decisions under risk and ambiguity (Sec.~\ref{sec:expr-human-choice}), OptScientist discovers an optimization law combining expected reward, cognitive switching cost and counterfactual reward. In central-bank monetary decisions (Sec.~\ref{sec:expr-central-bank}), it discovers a law in which observed policy tracks a latent subjective target through squared-error minimization. Across both domains, the discovered laws remain competitive with the strongest
baselines while using near-minimal parameter counts and symbolic complexity,
and substantially fewer output tokens than LLM-based model-discovery methods.

Overall, our contributions are threefold. First, we introduce the OptScientist Model, which, to the best of our knowledge, is the first scientific machine-learning framework for learning symbolic optimization problems directly from observational data. Second, for science, we develop OptScientist, a hybrid symbolic--LLM system that extends automated scientific discovery to optimization-law discovery in domains associated with optimization models. Third, at the paradigm level, we contribute to inverse optimization, revealed preference and related fields by moving beyond parameter estimation for pre-specified optimization problems, providing a new methodology for directly recovering optimization laws themselves from data.

\section{Related Work} \label{sec:literature}

\textbf{Symbolic discovery.}
Recent advances in symbolic regression have enabled symbolic law discovery from data. Most existing approaches, however, recover input--output laws, such as \(y=f(x)\) \citep{koza1994genetic,schmidt2009distilling,udrescu2020ai}, or physical equations expressed through differential relationships, such as \(\frac{d x}{d t} = f(x,t)\) \citep{brunton2016discovering,rudy2017data,champion2019data}. They are less suited to settings where the scientific law is itself an optimization problem. OptScientist learns symbolic optimization problems from observational data, providing a new symbolic-discovery model for scientific domains associated with optimization models.

\textbf{LLM-based automated scientific discovery.}
LLMs have been used to propose and refine symbolic functional forms or governing equations
\citep{merler2024context,du2024llm4ed,shojaee2025llm}, and to automate broader scientific workflows such as hypothesis generation, experimentation and iterative refinement
\citep{lu2026towards,gottweis2026accelerating,aygun2026ai}. Recent LLM-based work also searches for symbolic utility expressions to support group-level prediction \citep{zhao2026personalized}, but rarely recovers parsimonious scientific laws. Despite their broad role in scientific automation, these systems are not designed to discover compact symbolic optimization problems from data. OptScientist does not remove LLMs entirely. Instead, it suggests that greater efficiency and performance can be achieved by combining a symbolic discovery model with an auxiliary LLM agentic system to improve and interpret the discovered laws and add semantic meaning to them.

\textbf{Cognitive discovery.}
Some cognitive-discovery methods use black-box predictions or structured neural architectures to recover interpretable mechanisms from behavioural data
\citep{agrawal2020scaling,peterson2021using,miller2023cognitive,zhu2025capturing,ji2025discovering,ma2026behavior},
while others use LLMs to generate and revise executable models, diagnose theory failures and design informative experiments
\citep{binz2025automated,rmus2026generating,eltetHo2026atlas,jagadish2026closing}.
These approaches recover programs or reduced-form decision rules rather than compact symbolic optimization laws. Moreover, human behaviour is often modelled through resource-rational analysis \citep{debreu1959theory,lieder2020resource,zhu2026computation}, which formulates behaviour using hand-specified optimization models. As a general framework for automated discovery of optimization laws, OptScientist can support automated resource-rational analysis.

\textbf{Inverse optimization.}
Inverse optimization \citep{ahuja2001inverse,aswani2018inverse} and related approaches, including revealed preference \citep{afriat1967construction,varian1982nonparametric}, inverse reinforcement learning \citep{ng2000algorithms,ziebart2008maximum}, and optimization-based hierarchical network models \citep{ma2026behavior}, infer latent objectives or estimate the parameters of pre-specified optimization problems. However, they generally assume the functional form of the optimization problems. OptScientist provides a new methodology in which the optimization law itself can be recovered from data without prespecifying its functional form.

\section{The OptScientist} \label{sec:optsci}

\subsection{Formalizing Optimization-Law Discovery}

Generally, a law expressed as optimization describes an outcome $\mathbf{y}$ as the solution to an objective under constraints,
\begin{equation}
\max_{\mathbf{y} \in \mathcal{Y}}\,
U(\mathbf{x},\mathbf{y})
\quad\text{s.t.}\quad
\mathcal{C}(\mathbf{x},\mathbf{y}) \le 0,\;\;
\mathcal{T}(\mathbf{x},\mathbf{y}) = 0 .
\label{eq:ump}
\end{equation}
Here $\mathbf{x}$ denotes the observed context, $U(\cdot)$ is the objective, $\mathcal{C}(\cdot)$ represents inequality constraints, and $\mathcal{T}(\cdot)$ represents equality constraints or target conditions. This form covers a broad class of scientific laws in which observed outcomes arise from maximizing a quantity, possibly under constraints.

For learning, it is useful to express the same object as a single scalar landscape. Under standard local regularity conditions, the constrained problem in Eq.~\eqref{eq:ump} admits a locally exact penalty representation \citep{han1979exact},
\begin{equation}
\max_{\mathbf{y}\in\mathcal{Y}}\;
\lambda_0\,\phi\!\bigl(U(\mathbf{x},\mathbf{y})\bigr)
-\lambda_1^\top \rho\!\bigl(\mathcal{C}(\mathbf{x},\mathbf{y})\bigr)
-\lambda_2^\top \psi\!\bigl(\mathcal{T}(\mathbf{x},\mathbf{y})\bigr),
\label{eq:penalty-ump}
\end{equation}
Here $\phi$ is strictly increasing, $\rho(z)=\max\{z,0\}$ penalizes violated inequality constraints, and $\psi(z)=|z|$ penalizes deviations from equality constraints. This reformulation places objectives and constraints inside one optimization score, making the law easier to analyse and learn.

The outcome generated by an optimization law can be represented in two ways. In the deterministic case, the predicted outcome is \(\widehat{\mathbf{y}}(\mathbf{x})=\arg\max_{\mathbf{y}\in\mathcal{Y}} S(\mathbf{x},\mathbf{y})\), where \(S(\mathbf{x},\mathbf{y})\) is a scalar representation of the underlying optimization problem. For example, under the exact-penalty reformulation in Eq.~\eqref{eq:penalty-ump}, \(S\) is given by the objective together with penalties for violated constraints.
In noisy systems, the same score defines a probability distribution over possible outcomes,
\begin{equation}
p(\mathbf{y}\mid\mathbf{x})
=
\exp\!\left(S(\mathbf{x},\mathbf{y})/\tau\right) / Z(\mathbf{x}),
\label{eq:gibbs-law}
\end{equation}
where $\tau>0$ controls the level of noise and $Z(\mathbf{x})$ is the normalizing term. As $\tau$ decreases, the distribution concentrates on the maximizers of $S$; for larger $\tau$, it describes noisy realizations of the same underlying optimization principle.

Therefore, we formulate the central question of optimization-law discovery as follows.

\begin{question}[Optimization-Law Discovery]
Given observational data
$\mathcal D=\{(\mathbf{x}_i,\mathbf{y}_i)\}_{i=1}^N$
generated by an unknown optimization law
$\mathcal L^*$ of the form
\[
\max_{\mathbf{y}\in\mathcal Y}
U^*(\mathbf{x},\mathbf{y})
\quad\text{s.t.}\quad
\mathcal C^*(\mathbf{x},\mathbf{y})\le 0,\;\;
\mathcal T^*(\mathbf{x},\mathbf{y})=0,
\]
where outcomes arise either as optimal solutions or
as stochastic realizations of the induced optimization
landscape, can we recover a compact symbolic law
\[
\widehat{\mathcal L}
=(\widehat U,\widehat{\mathcal C},\widehat{\mathcal T})
\]
without pre-specifying the functional forms of its
components, such that it reproduces the observable
implications of $\mathcal L^*$?
\end{question}


\begin{figure}[t]
    \centering
    \includegraphics[width=0.95\linewidth]{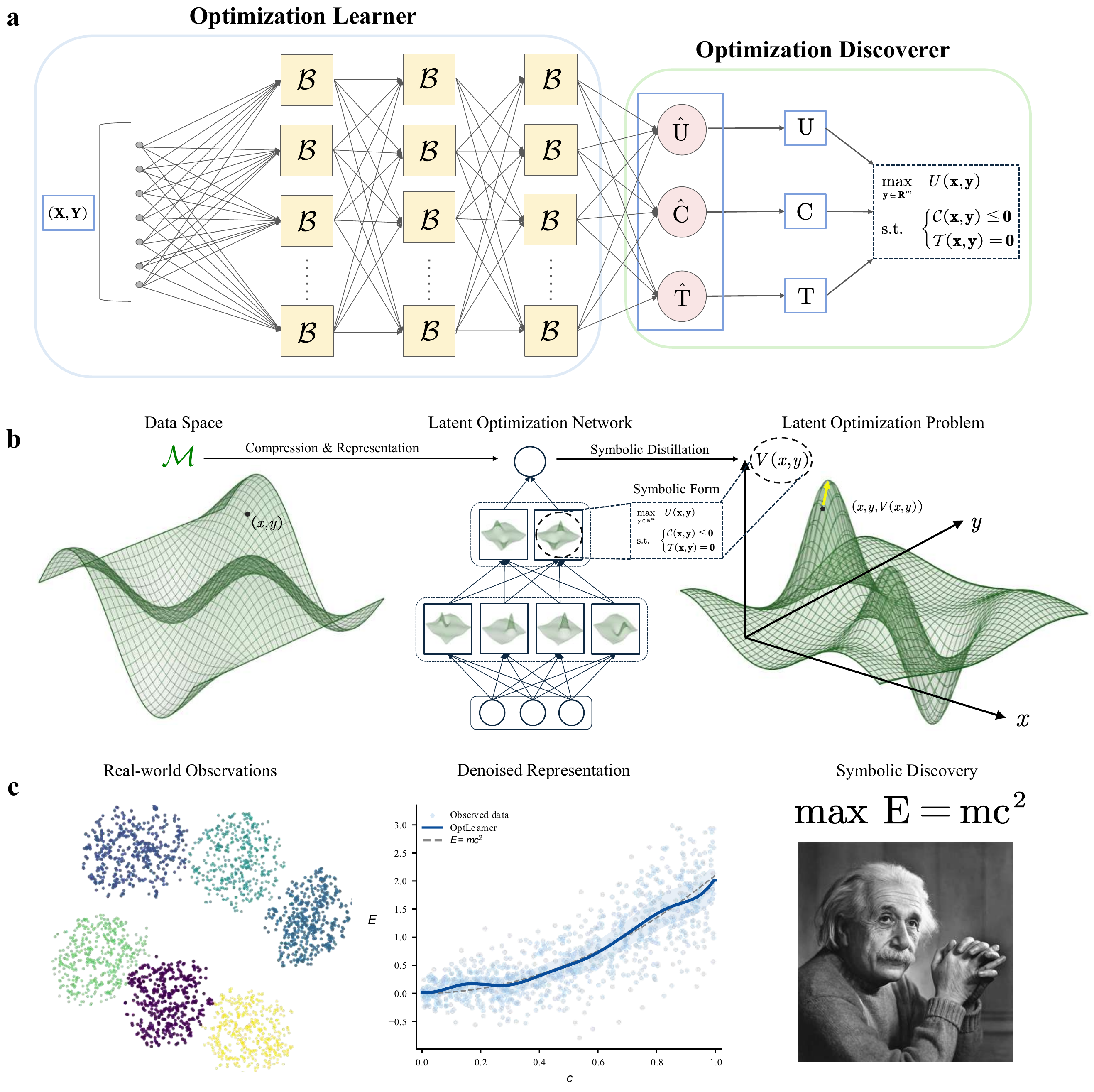}
    \caption{
    \textbf{The OptScientist Model.}
    \textbf{a,} The OptScientist Model consists of an Optimization Learner and an Optimization Discoverer. The Optimization Learner learns to represent observational data as an optimization problem using a hierarchical network with an optimization representation bottleneck. The Optimization Discoverer then distills this learned optimization representation into a compact symbolic optimization problem.
    \textbf{b,} The OptScientist Model represents the data manifold through a latent optimization network. Raw observations are first compressed and represented by hierarchical optimization blocks. The learned representations are then expressed as a single optimization problem through the optimization representation bottleneck, from which a compact symbolic law is extracted.
    \textbf{c,} The model follows the same logic in concrete discovery settings. Noisy real-world observations are first transformed into denoised representations, and symbolic regression is then applied to the learned optimization representation to recover symbolic laws.
    }
    \label{fig:optscientist}
\end{figure}

\subsection{OptScientist Model}
\label{sec:optscientist-model}

To discover compact optimization laws from data directly, the OptScientist Model consists of two stages: the Optimization Learner (OptLearner; see Figure \ref{fig:optscientist}a Left) and the Optimization Discoverer (OptDiscover; see Figure \ref{fig:optscientist}a Right).

\subsubsection{Optimization Learner}

Observational data rarely reveal the optimization law directly. They may be noisy, high-dimensional, redundant, or expressed in variables that are not yet suitable for symbolic modelling. 
The Optimization Learner (OptLearner) addresses this challenge by learning to represent observational data as an optimization problem through a hierarchical optimization network with an optimization representation bottleneck. This process enables denoising, information compression, and automatic feature extraction, producing a learned optimization representation for subsequent symbolic discovery.

\textbf{Behavior Learning backbone.}
OptLearner uses the Behavior Learning Network (BL) \citep{ma2026behavior, ma2026arxivbehavior} as its backbone to learn a scalar landscape over context--outcome pairs. For each context $\mathbf{x}$ and candidate outcome $\mathbf{y}$, the network returns a score $S_{\theta}(\mathbf{x},\mathbf{y})$ that measures how strongly $\mathbf{y}$ is preferred under the latent optimization problem. Under Eq.~\eqref{eq:gibbs-law}, observed outcomes are treated as stochastic samples from the learned optimization landscape. Learning therefore reconstructs the latent landscape on which observed outcomes arise as preferred solutions. This corresponds to the representation-learning step in Figure~\ref{fig:optscientist} (a,b).

\textbf{Optimization representation bottleneck.}
OptLearner builds on the BL backbone to represent observational data as a single optimization problem. Its early layers denoise and compress the observations, capturing interactions and latent features that may be difficult to express in the original coordinates. The final layer consists of a single BL block, forming an optimization representation bottleneck with learnable objective and optional constraint components, each potentially comprising multiple branches,
\begin{equation}
\max_{\mathbf{y} \in \mathcal{Y}}\,
\underbrace{
U(\mathbf{x},\mathbf{y})
}_{[\lambda_0]}
\quad\text{s.t.}\quad
\underbrace{
\mathcal{C}(\mathbf{x},\mathbf{y})
}_{[\boldsymbol{\lambda}_1]}
\le 0,\;\;
\underbrace{
\mathcal{T}(\mathbf{x},\mathbf{y})
}_{[\boldsymbol{\lambda}_2]}
= 0 .
\label{eq:single-bl-block}
\end{equation}
Here \(\lambda_0\) controls the scale of the objective, while
\(\boldsymbol{\lambda}_1\) and \(\boldsymbol{\lambda}_2\) control the strengths of the inequality and equality terms. This bottleneck provides a structured optimization representation for subsequent symbolic discovery. When the input variables are already meaningful, OptLearner primarily denoises the optimization landscape. For high-dimensional or unstructured inputs, it additionally extracts and compresses relevant features before forming the learned optimization representation.

\textbf{Training objective.}
The training objective follows the same principle for discrete and continuous outcomes. For a finite outcome set $\mathcal{Y}$, OptLearner evaluates all candidates and trains the score by cross-entropy,
\begin{equation}
\mathcal{L}_{\mathrm{disc}}
=
-\frac{1}{N}\sum_{i=1}^{N}
\log
\frac{\exp\!\left(S_{\theta}(\mathbf{x}_i,\mathbf{y}_i)/\tau\right)}
{\sum_{\mathbf{y}\in\mathcal{Y}}
\exp\!\left(S_{\theta}(\mathbf{x}_i,\mathbf{y})/\tau\right)} .
\label{eq:discrete-loss}
\end{equation}
For continuous outcomes, enumerating $\mathcal{Y}$ is infeasible. OptLearner therefore compares the observed outcome with sampled alternatives $\widetilde{\mathbf{y}}_{ik}$
using the same contrastive objective,
\begin{equation}
\mathcal{L}_{\mathrm{cont}}
=
-\frac{1}{N}\sum_{i=1}^{N}
\log
\frac{\exp\!\left(S_{\theta}(\mathbf{x}_i,\mathbf{y}_i)/\tau\right)}
{\exp\!\left(S_{\theta}(\mathbf{x}_i,\mathbf{y}_i)/\tau\right)
+
\sum_{k=1}^{K}
\exp\!\left(S_{\theta}(\mathbf{x}_i,\widetilde{\mathbf{y}}_{ik})/\tau\right)} .
\label{eq:continuous-loss}
\end{equation}
Both losses encode the same principle: under the observed context, the observed outcome should receive the highest optimization score.


\subsubsection{Optimization Discoverer}
\label{sec:optdiscover}

The Optimization Learner represents observational data as an optimization problem, but the learned representation is not yet a compact symbolic law. The Optimization Discoverer (OptDiscover) addresses this by extracting the symbolic form of the learned optimization problem, as illustrated in Figure~\ref{fig:optscientist} (a,b). After OptLearner is trained, its parameters are frozen. OptDiscover then distills the components of the optimization representation bottleneck into compact symbolic expressions. This bottleneck consists of three types of components: objective terms \(U_{\theta}\), inequality-constraint terms \(C_{\theta}\), and equality-constraint terms \(T_{\theta}\), with each type optionally comprising multiple branches. OptDiscover searches for a compact symbolic expression for each active component,
\begin{equation}
\widehat{H}_{j}
=
\arg\min_{F\in\mathcal{F}_{\Omega_j}}
\frac{1}{M}\sum_{m=1}^{M}
\left[
F(\mathbf{x}_m,\mathbf{y}_m)
-
H_{\theta,j}(\mathbf{x}_m,\mathbf{y}_m)
\right]^2 ,
\qquad
H_{\theta,j}\in\{U_{\theta},C_{\theta},T_{\theta}\}.
\label{eq:term-wise-symbolic-distillation}
\end{equation}
Here \(\mathcal{F}_{\Omega_j}\) denotes the symbolic search space for the \(j\)-th term. 
The distilled expressions of the active components are assembled according to the learned optimization structure, yielding a symbolic optimization problem with explicit objectives, inequality constraints, and equality constraints.

\textbf{Optional automatic feature extraction.}
OptDiscover can operate with or without automatic feature extraction. When the observed variables already form a meaningful symbolic language, the law is distilled directly from \((\mathbf{x},\mathbf{y})\). When the raw variables require transformation, interaction, or compression, OptLearner first maps them into denoised intermediate representations. Technically, these representations are the outputs of intermediate layers in OptLearner, taken before the optimization representation bottleneck. These layer-level variables act as automatically extracted features, and OptDiscover then extracts the symbolic law on this learned feature space. This allows the framework to handle both meaningful scientific variables and raw observations whose symbolic structure is more naturally expressed after representation learning.

\textbf{Complexity penalty.}
Symbolic complexity is measured by expression-tree size, with each
variable, numerical constant and operator counted as one node. OptDiscover controls symbolic complexity in two ways. First, it selects the simplest expression among candidates with near-minimal distillation loss. This favours parsimonious laws while preserving the learned optimization structure. Second, for the objective term, OptDiscover removes additive components that depend only on \(\mathbf{x}\). These components are constant across candidate outcomes \(\mathbf{y}\) and therefore do not affect the optimizer.


\subsection{OptScientist Agent}
\label{sec:optscientist-agent}

\begin{wrapfigure}{r}{0.55\textwidth}
    \centering
    \includegraphics[width=0.53\textwidth]{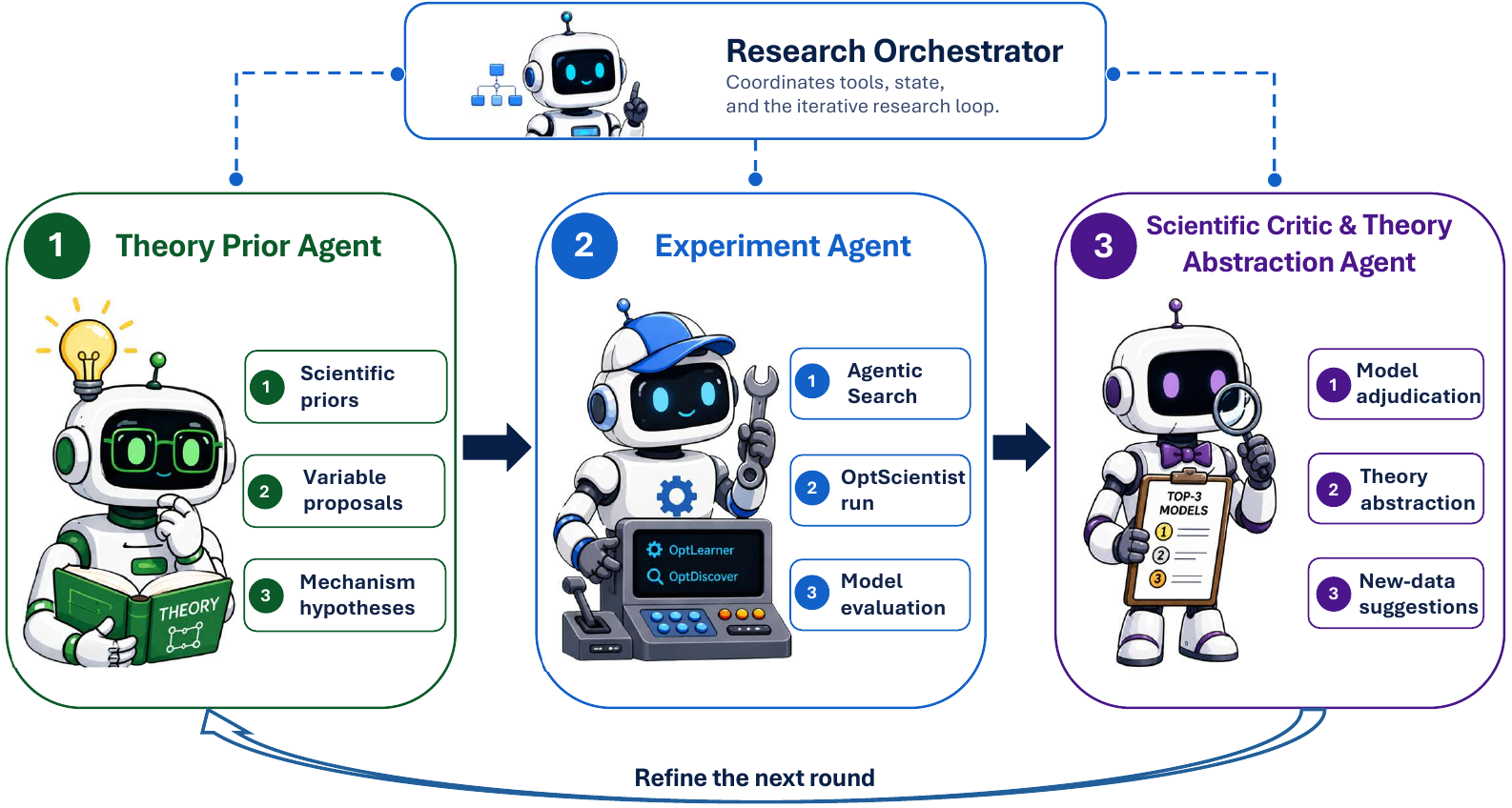}
    \caption{
    \textbf{OptScientist Agent as an iterative research framework.}
    The framework embeds the OptScientist Model in a scientific-discovery loop. 
    }
    \label{fig:optscientist-agent}
\end{wrapfigure}


Symbolic discovery often requires scientific decisions both before and after model fitting: researchers construct meaningful variables, impose search priors, interpret candidate expressions, abstract qualitative mechanisms and decide how subsequent analysis should be revised. To automate this research loop, OptScientist Agent embeds the OptScientist Model in an iterative framework for optimization-law discovery (Figure~\ref{fig:optscientist-agent}). The system comprises four interacting modules: a Research Orchestrator, a Theory Prior Agent, an Experiment Agent and a Scientific Critic and Theory Abstraction Agent. A detailed description of the OptScientist Agent is provided in Appendix~\ref{sec:add-optscientist-agent}.

\textbf{Research Orchestrator.}
The Research Orchestrator maintains the shared research state and coordinates the discovery process. It integrates hypotheses, candidate variables, discovered laws and scientific findings across agents and discovery rounds, while carrying unresolved questions forward. It also determines whether the workflow should terminate or enter another round of refinement, turning OptScientist runs into an iterative scientific-discovery process.

\textbf{Theory Prior Agent.}
The Theory Prior Agent draws on the scientific literature to identify established theories and mechanisms relevant to the research question, and translates them into explicit, testable hypotheses. It then proposes scientifically meaningful variables or transformations that represent these mechanisms, together with symbolic-search priors when scientifically justified. These priors guide the search while keeping the final expression data-driven.

\textbf{Experiment Agent.}
Guided by the priors from the Theory Prior Agent, the Experiment Agent translates scientific hypotheses, candidate variables and search priors into executable OptScientist configurations. It then uses the core OptScientist Model to search for candidate optimization laws from observational data. The resulting laws and supporting empirical evidence are returned for scientific critique and theory abstraction.

\textbf{Scientific Critic and Theory Abstraction Agent.}
The Scientific Critic and Theory Abstraction Agent evaluates candidate laws as scientific explanations and identifies the best-supported structures across them. It synthesizes the supported laws into a more general theoretical form and summarizes their qualitative mechanisms and main scientific findings. Remaining uncertainty is then translated into new hypotheses, variables or data requirements for the next discovery round, moving OptScientist Agent from symbolic law recovery towards iterative theory refinement.

\textbf{Embedding the OptScientist Model in the discovery loop.}
The OptScientist Model forms the empirical core of the agentic system. In each round, scientific hypotheses, candidate variables and search priors are translated into executable configurations, through which the model discovers candidate optimization laws from observational data. The resulting laws and empirical evidence are then returned to the agents for evaluation, theory abstraction and refinement of the next round.


\section{Experiments} \label{sec:expr}

We evaluate OptScientist Model on \emph{OptFeynman}, a new benchmark for optimization-law discovery, and the full framework on two real-world scientific-discovery tasks, with comparisons against machine-learning models, symbolic regression, and LLM model-discovery methods. The LLM baselines include an LLM-only researcher, GeCCo \citep{rmus2026generating} as a code-assisted LLM approach, and Model Discovery Agent \citep[MDA;][]{murphy2026model} as a Bayesian experiment-design approach. All LLM methods use DeepSeek-V4-Flash, the same LLM used by the optional OptScientist Agent. Data and evaluation are provided in Sec.~\ref{sec:data}, and comparison methods in Appendix~\ref{sec:baselines}.


\begin{figure}[htbp]
    \centering
    \includegraphics[width=\textwidth]{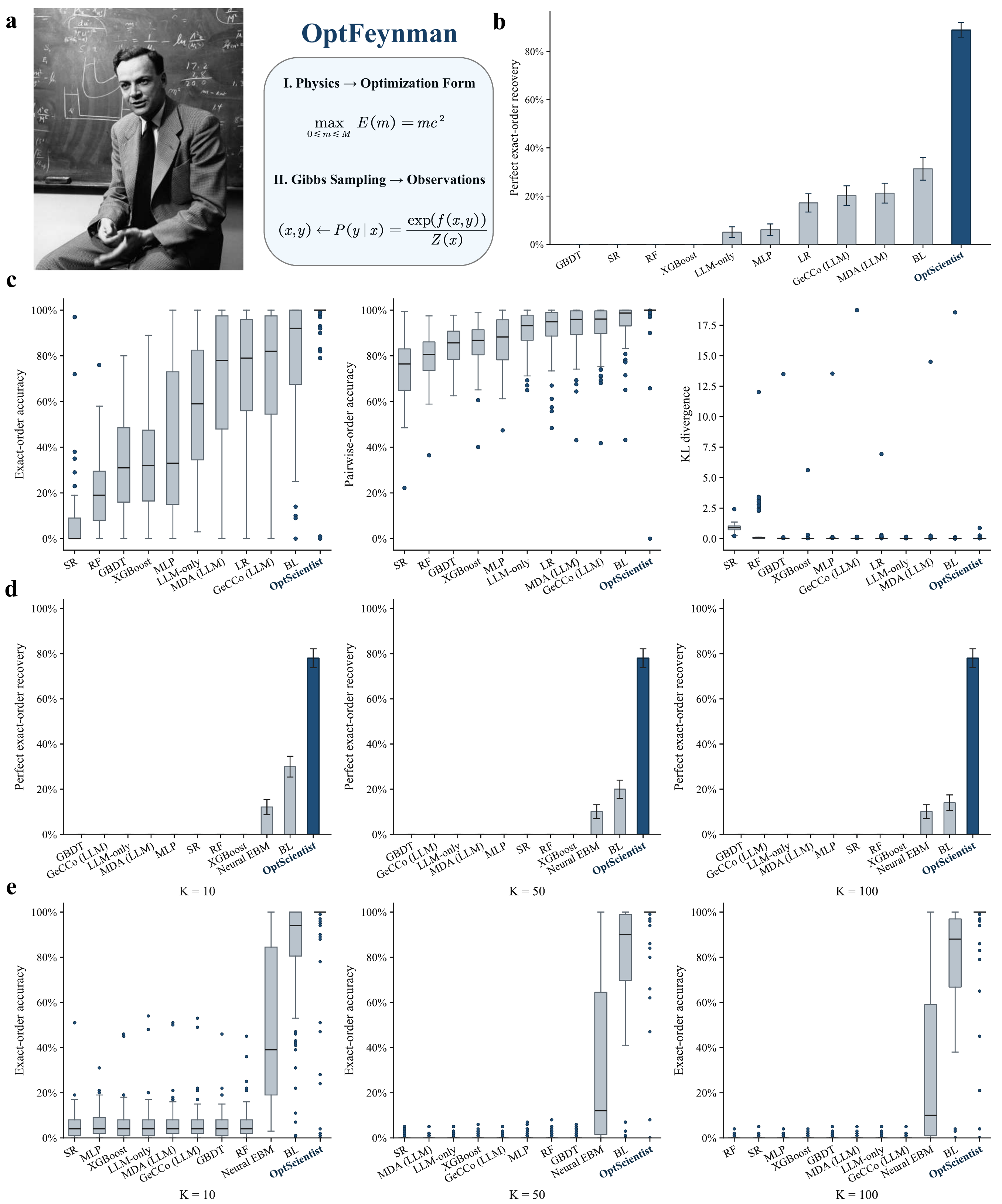}
    \caption{
    \textbf{Recovering ground-truth optimization laws on OptFeynman.}
    \textbf{a,} OptFeynman converts analytic physics equations into latent
    optimization problems and generates observations by Gibbs sampling from the
    induced landscapes.
    \textbf{b,c,} For discrete decision variables, OptScientist shows the
    strongest out-of-sample ability to recover the true optimization law on
    unseen test contexts, outperforming baselines. It achieves perfect exact-order recovery on
    \(88/99\) tasks, indicating that the recovered symbolic laws induce
    preference orders approximately equivalent to the ground-truth laws.
    \textbf{d,e,} For continuous decision variables, the advantage persists on
    unseen test contexts across increasingly dense candidate sets. OptScientist
    perfectly recovers the exact order on \(78/99\) tasks at \(K=10\), \(K=50\)
    and \(K=100\), showing robust discovery of the underlying optimization law
    beyond discrete-choice settings. Error bars denote standard errors across datasets where
    shown.
    }
    \label{fig:optfeynman-standard}
\end{figure}

\begin{figure}[t]
    \centering
    \includegraphics[width=\textwidth]{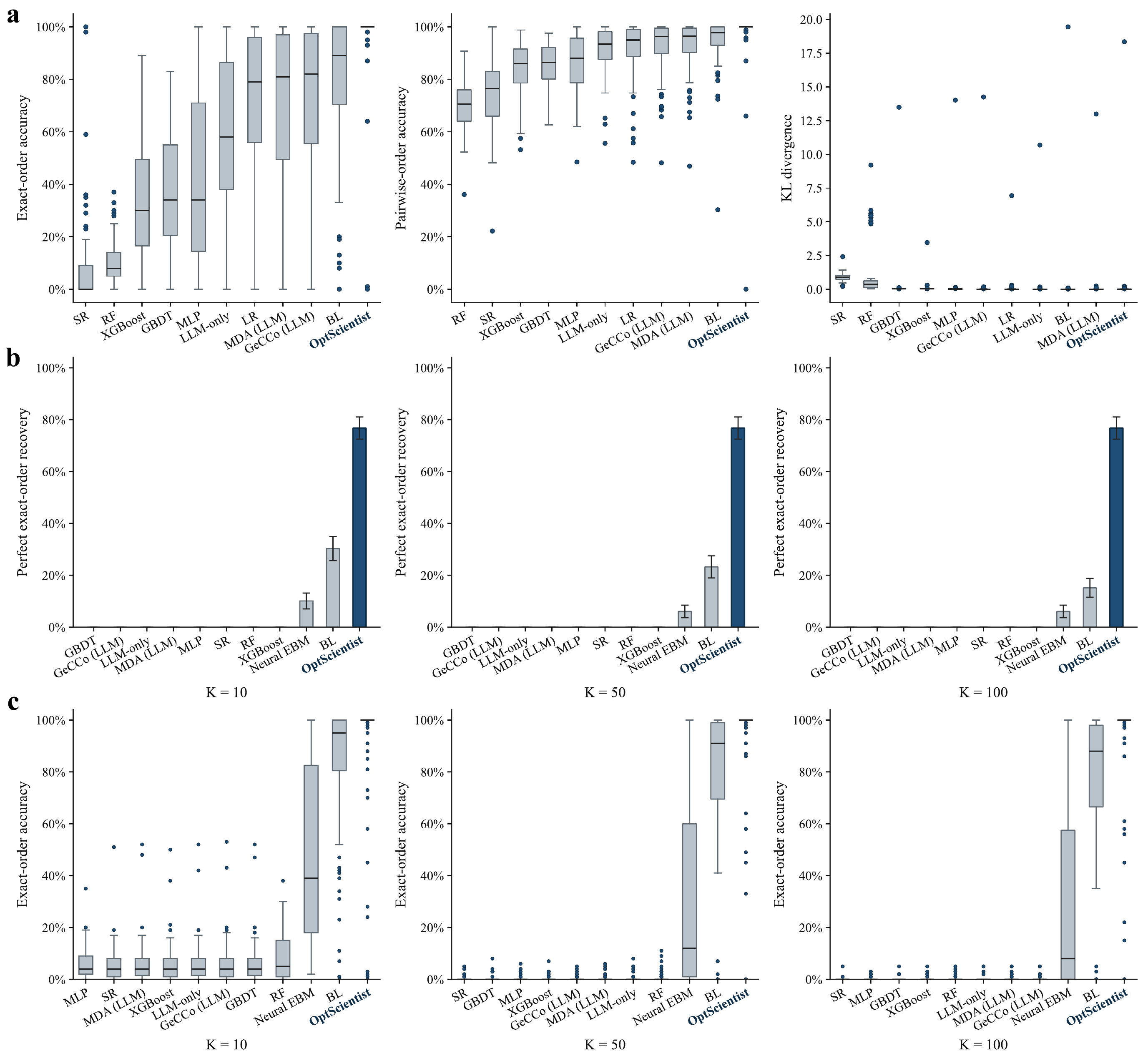}
    \caption{
    \textbf{Robust recovery of optimization laws under noisy context variables.}
    Gaussian noise is added to the training context variables, while validation
    and test contexts remain clean.
    \textbf{a,} For discrete decision variables, OptScientist remains the
    strongest method for out-of-sample recovery of the true optimization law,
    achieving perfect exact-order recovery on \(92/99\) tasks and outperforming
    machine-learning, symbolic-regression, optimization-structured, and LLM-based automated model-discovery baselines.
    \textbf{b,c,} For continuous decision variables, OptScientist retains high
    recovery performance on unseen test contexts across \(K=10\), \(K=50\) and
    \(K=100\), perfectly recovering the exact order on \(76/99\) tasks at each
    candidate-set size and continuing to outperform the evaluated baselines. These results show that the model can recover
    approximately equivalent ground-truth optimization laws even when the
    training observations are corrupted by context noise. Error bars denote standard errors across
    datasets where shown.
    }
    \label{fig:optfeynman-noisyx}
\end{figure}

\subsection{Recovering Ground-Truth Optimization Laws}
\label{sec:expr-ground-truth}

\begin{wraptable}{r}{0.50\textwidth}
\vspace{-20pt}
\centering
\tiny
\setlength{\tabcolsep}{3pt}
\renewcommand{\arraystretch}{1.25}
\caption{
\textbf{Examples of recovered OptFeynman laws.}
Examples show that OptScientist perfectly recovers effective ground-truth optimization expressions across diverse physical structures. 
} \vspace{2mm}
\label{tab:optfeynman-recovered-examples-tiny}
\begin{tabular}{@{}p{0.20\linewidth}p{0.35\linewidth}p{0.35\linewidth}@{}}
\toprule
Decision variable & Effective ground truth & Effective recovered expression \\
\midrule
\(y_3\)
& \(\displaystyle \max_{y_3}\; x_3 y_3\)
& \(\displaystyle \max_{y_3}\; x_3 y_3\) \\
\addlinespace[0.35em]

\(w\)
& \(\displaystyle \max_{w}\; w^2\)
& \(\displaystyle \max_{w}\; w^2\) \\
\addlinespace[0.35em]

\(E_f\)
& \(\displaystyle \max_{E_f}\; E_f\cos\theta\)
& \(\displaystyle \max_{E_f}\; E_f\cos\theta\) \\
\addlinespace[0.35em]

\(G\)
& \(\displaystyle \max_{G}\; G\left(\frac{1}{r_2}-\frac{1}{r_1}\right)\)
& \(\displaystyle \max_{G}\; \frac{G(r_1-r_2)}{r_1r_2}\) \\
\addlinespace[0.35em]

\(d\)
& \(\displaystyle \max_{d}\; \frac{T_2-T_1}{d}\)
& \(\displaystyle \max_{d}\; \frac{T_2-T_1}{d}\) \\
\addlinespace[0.35em]

\(r\)
& \(\displaystyle \max_{r}\; \frac{1}{r^2}\)
& \(\displaystyle \max_{r}\; \frac{1}{r^2}\) \\
\addlinespace[0.35em]

\(d\)
& \(\displaystyle \max_{d}\; \frac{1}{d}\)
& \(\displaystyle \max_{d}\; \frac{1}{d}\) \\
\bottomrule
\end{tabular}
\vspace{-18pt}
\end{wraptable}

We first tested whether OptScientist Model can recover ground-truth optimization laws
from observations. OptFeynman converts analytic physics equations into latent optimization problems and generates observations via Gibbs sampling. Because
the generating objective is known, recovery can be evaluated directly on unseen
test contexts by comparing the preference order and Gibbs distribution of the recovered
and ground-truth laws. \textit{Notably, this experiment evaluates only the OptScientist Model, without any LLM assistance.}

\medskip
\noindent\textbf{Ground-truth recovery on OptFeynman.}
OptScientist shows the strongest out-of-sample ability to recover the true
optimization law across both discrete and continuous decision variables
(Figure~\ref{fig:optfeynman-standard}). For discrete decision variables, it achieves perfect exact-order recovery on \(88/99\) problems, indicating that the recovered symbolic laws induce preference orders
approximately equivalent to the ground-truth laws. For continuous decision
variables, OptScientist achieves perfect exact-order recovery on \(78/99\) problems
at \(K=10, 50, \text{~and~} 100\). In both settings, OptScientist substantially outperforms all evaluated baselines. The stability of performance across increasingly
dense candidate sets shows robust discovery of the underlying law.
The recovered symbolic laws also preserve the stochastic structure of the
generating landscape, yielding Gibbs distributions close to the ground truth.

\textbf{Robustness to noisy context variables.}
We next evaluated whether recovery remains reliable when the training
observations are corrupted by context noise (Figure~\ref{fig:optfeynman-noisyx}).
Specifically, Gaussian noise is added to the training context variables, while validation and
test contexts remain clean. Under this setting, OptScientist remains the
strongest method for out-of-sample recovery of the true optimization law. It
achieves perfect exact-order recovery on \(92/99\) discrete problems and on
\(76/99\) continuous problems at \(K=10\), \(K=50\) and \(K=100\). These results
show that OptScientist can recover approximately equivalent ground-truth
optimization laws even when the training observations are corrupted by context
noise.

\textbf{Representative recovered laws.}
Table~\ref{tab:optfeynman-recovered-examples-tiny} shows representative examples of
successful symbolic recovery. The comparison focuses on the decision-relevant
part of each optimization law after pruning numerical constants and scale
factors that do not affect the optimizer. OptScientist recovers diverse physical
structures, including linear interactions, quadratic energy terms,
inverse-square laws, heat-transfer terms and algebraically equivalent rational
forms. These examples illustrate that the recovered laws preserve the
optimization law up to choice-equivalent symbolic transformations.




\subsection{Human Choice under Risk and Ambiguity}
\label{sec:expr-human-choice}

\begin{figure}[htbp]
    \centering
    \includegraphics[width=\textwidth]{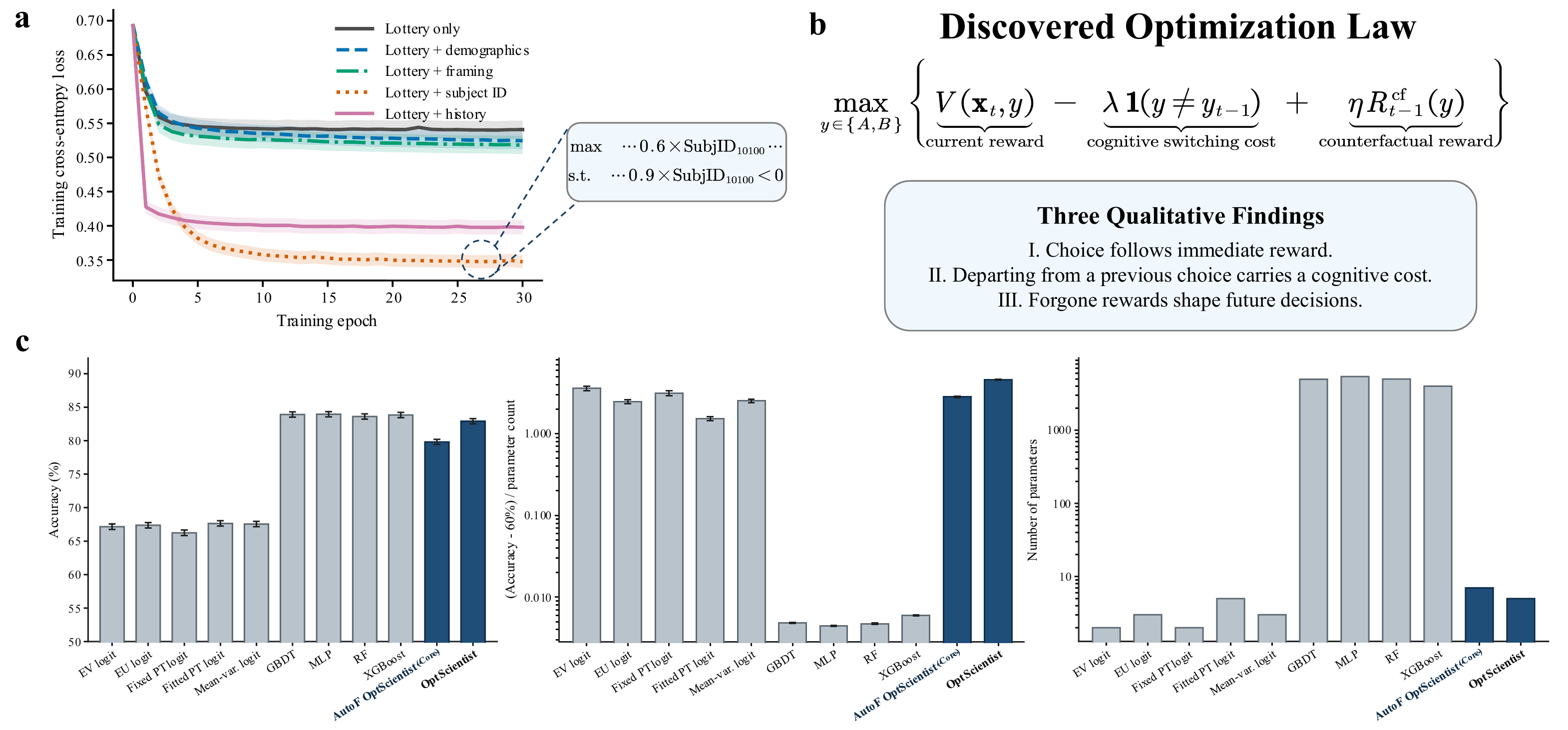}
    \caption{
    \textbf{A compact law of human decision-making under risk and ambiguity.}
    \textbf{a,} A preliminary analysis of the CPC18 repeated-choice task shows that
decision-history variables are particularly informative relative to lottery,
demographic and framing variables, consistent with the sequential structure of
the discovered optimization law.
    \textbf{b,} OptScientist recovers this sequential structure as a compact
    symbolic law: people tend to follow immediate reward, resist switching from
    a previous choice and remain sensitive to forgone outcomes.
    \textbf{c,} Across behavioural baselines, machine-learning models and
    OptScientist variants, the recovered law achieves the near-best predictive
    performance with a near-minimal number of parameters. The result shows that
    human choice can be compressed into a simple optimization principle rather
    than explained by model scale alone. Error bars denote the standard error
    of the mean across participant-level metrics.
    }
    \label{fig:cpc18}
\end{figure}

We next applied OptScientist to human decisions under risk and ambiguity, where
the underlying optimization law is unknown and human choices are shaped by
repeated feedback, individual heterogeneity and noisy behavioural responses.
The CPC18 dataset asks participants to repeatedly choose between two monetary
prospects \citep{plonsky2019cpc18}. This setting tests how OptScientist turns 
behavioural data into a compact symbolic optimization law.

\textbf{Theory-guided generation of supplementary variables.}
Guided by scientific literature, the Theory Prior Agent constructs theoretically motivated variables beyond those in the original data, including expected value, forgone payoffs, gender and residence in wealthier regions. These variables are then considered in symbolic discovery. See Appendix~\ref{sec:add-cpc18-prior-analysis} for details.


\textbf{A compact law of repeated choice.}
OptScientist recovered a compact symbolic optimization law
(Figure~\ref{fig:cpc18}b). 
In abstract form, the discovered law is
\begin{equation}
    \max_{y_t\in\{A,B\}}
    \left\{
    V(x_t,y_t)
    - \lambda\,\mathbf{1}(y_t\neq y_{t-1})
    + \eta\,R^{\mathrm{cf}}_{t-1}(y_t)
    \right\}.
    \label{eq:cpc18-abstract-law}
\end{equation}
Here, \(V(x_t,y_t)\) is the current reward value of the candidate option,
\(\mathbf{1}(y_t\neq y_{t-1})\) captures the cognitive cost of switching from
the previous choice, and \(R^{\mathrm{cf}}_{t-1}(y_t)\) captures sensitivity to
the forgone reward from the previous round. This law summarizes three
qualitative mechanisms: choice follows immediate reward, departing from a
previous choice carries a cognitive cost, and forgone rewards shape future
decisions. In this case, our approach can be viewed as performing an automated resource-rational analysis of human risky choices. The concrete score-difference expression and its mapping to this
abstract law are given in Additional Experimental Details,
Appendix~\ref{sec:add-cpc18-direct-law}.


\textbf{Automatic-feature symbolic discovery.}
Following the feature-mediated design introduced in the OptScientist Model, we
also evaluated an automatic-feature variant, denoted AutoF OptScientist. In this
variant, OptLearner first maps \((x_t,y_t)\) into denoised intermediate
representations from its inner layers. These layer-level variables act as
automatically extracted features, and OptDiscover then distills the final
optimization score as a symbolic function on the learned feature space. This
provides an additional compact symbolic description of the CPC18 dataset and
allows representation learning to support symbolic discovery when behavioural
variables are noisy, redundant or weakly structured. See Appendix~\ref{sec:add-cpc18-autof-law} for the recovered inner-feature expression.


\textbf{Near-best performance with near-minimal complexity.}
Across behavioural baselines, machine-learning models and OptScientist
variants, the recovered law achieves a favourable accuracy--complexity trade-off
(Figure~\ref{fig:cpc18}c). The direct OptScientist law reaches \(82.9\%\) accuracy
with five numerical constants, approaching the strongest high-capacity
machine-learning baselines that use thousands of parameters. Notably, predictive performance on CPC18 appears to saturate at approximately
83\% held-out accuracy across strong predictive models. AutoF OptScientist
achieves comparable ranking and calibration performance with seven numerical
constants, including an AUC of \(0.865\), a probability MSE of \(0.137\), and an
ECE of \(0.116\). Full predictive metrics and complexity definitions are
reported in Additional Experimental Details,
Appendix~\ref{sec:add-cpc18-performance}. These results show that repeated human
choice in CPC18 can be compressed into a simple optimization principle
combining immediate reward, cognitive switching cost and counterfactual reward.

\subsection{Central-Bank Monetary Policy}
\label{sec:expr-central-bank}

\begin{figure}[htbp]
    \centering
\includegraphics[width=\textwidth]{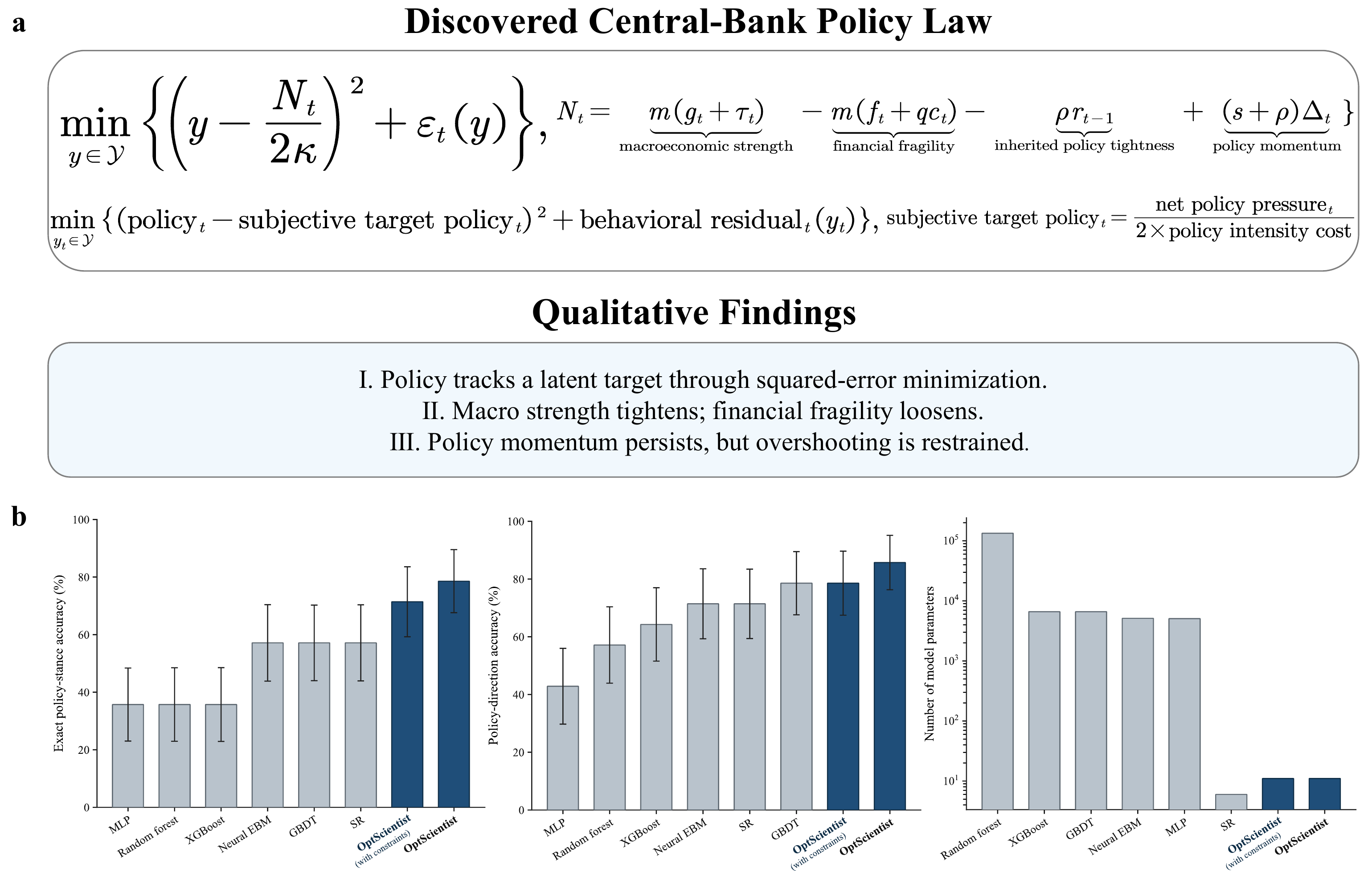}
\caption{
\textbf{Discovery of a central-bank policy law.}
\textbf{a,} OptScientist recovers a compact symbolic rule for quarterly
monetary-policy decisions. The five-class policy stance, ranging from large
easing to large tightening, is explained as tracking a latent policy target
driven by macroeconomic strength, financial fragility, inherited tightness and
policy momentum.
\textbf{b,} On a future-period out-of-time test set, OptScientist attains the best policy-stance and policy-direction
accuracy while using a near-minimal parameter count. Error bars denote bootstrap standard errors from \(10{,}000\) quarter-level
resamples with replacement.
}
    \label{fig:central-bank}
\end{figure}

We then applied OptScientist to quarterly central-bank monetary-policy
decisions, where the observed action is a five-class policy stance ranging from
large easing to large tightening. The task is to discover a compact
optimization law linking macro-financial conditions to policy actions. This
setting tests how OptScientist recovers a symbolic policy rule from
small, noisy and temporally structured macroeconomic data.

\textbf{Discovery of a central-bank optimization law.}
OptScientist recovers a compact symbolic optimization law for monetary-policy
decisions. For scientific interpretation, the selected law is abstracted into a
policy-target tracking form (see Figure~\ref{fig:central-bank}a),
\begin{equation}
    \min_{y_t\in\mathcal{Y}}
    \left\{
    \left(
    y_t-\frac{N_t}{2\kappa}
    \right)^2
    +\varepsilon_t(y_t)
    \right\},
    \label{eq:central-bank-main-law}
\end{equation}
where \(y_t\) is the policy stance, \(\varepsilon_t(y_t)\) is a residual
decision term and \(N_t/(2\kappa)\) is the latent subjective target policy. The
recovered net policy pressure is \(N_t=m(g_t+\tau_t)-m(f_t+q c_t)-\rho r_{t-1}+(s+\rho)\Delta_t\). The first term captures macroeconomic strength, the second captures financial
fragility, the third captures inherited policy tightness and the fourth captures
policy momentum. The law therefore explains quarterly policy as tracking a
latent target: macroeconomic strength tightens policy, financial fragility
loosens policy, inherited tightness restrains further tightening, and policy
momentum carries previous adjustments forward. The full symbolic expression and
the algebraic rewriting into the target-tracking form are given in Additional
Experimental Details, Appendix~\ref{sec:add-central-bank-single-u}
and~\ref{sec:add-central-bank-target-rewrite}.



\textbf{Constrained policy-law decomposition.}
OptScientist also recovers a constrained symbolic law with explicit \(U\), \(C\) and \(T\) components, decomposing the policy score into policy pressure, a one-sided constraint and a quadratic risk or target-deviation term while retaining competitive predictive performance. The full expression and branch-level complexities are reported in Appendix~\ref{sec:add-central-bank-uct}.


\textbf{Near-best performance with near-minimal complexity.}
Across baselines and OptScientist variants, the single-\(U\) law achieves the
best point estimates for exact and direction accuracy
(Figure~\ref{fig:central-bank}b), reaching \(78.6\%\) and \(85.7\%\),
respectively, on the out-of-time test set with only 11 numerical parameters.
The constrained variant remains competitive at \(71.4\%\) and \(78.6\%\),
respectively, with the same parameter count. Full metrics and complexity details are reported in
Appendix~\ref{sec:add-central-bank-performance}.

\subsection{Token-Efficient Model Discovery}
\label{sec:token_efficiency}

\begin{figure}[htbp]
    \centering
    \includegraphics[width=\linewidth]{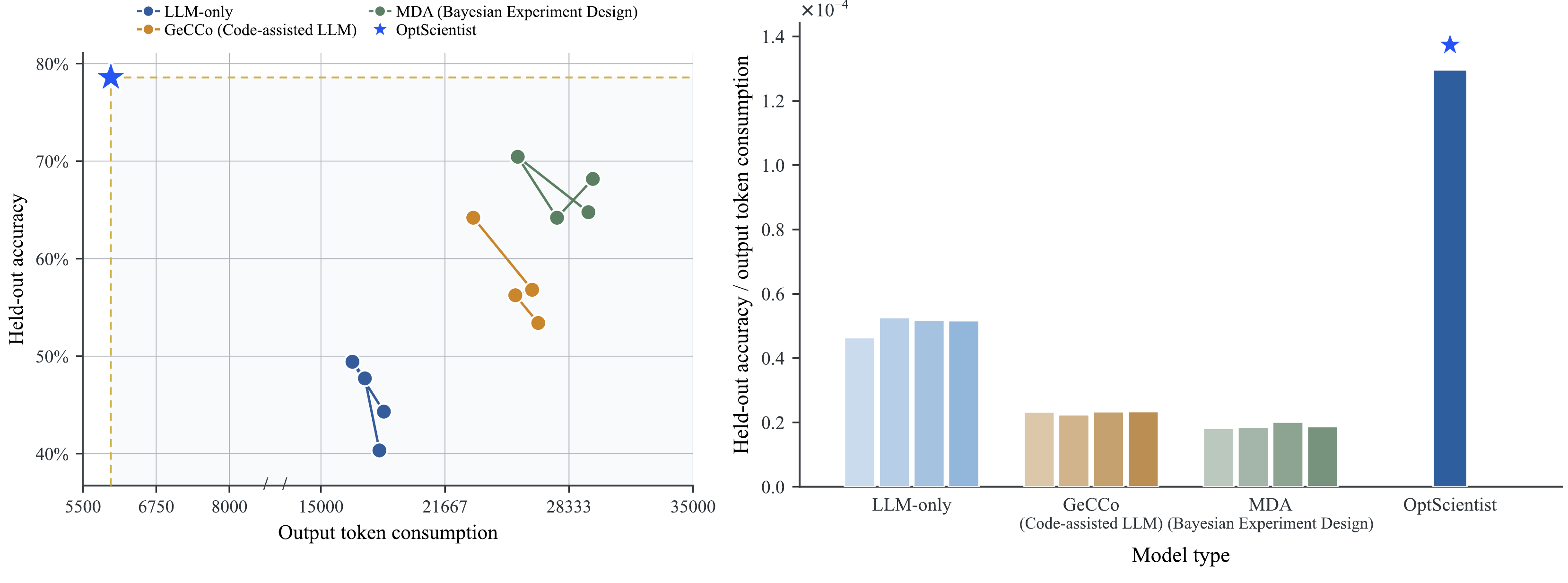}
    \caption{
    \textbf{Token-efficient automated model discovery.}
    We further compare the token efficiency of OptScientist with three LLM-based automated model-discovery methods on the two real-world discovery tasks: LLM-only discovery, GeCCo (code-assisted LLM), and MDA (Bayesian experiment design). All LLM components use DeepSeek-V4-Flash.
    \textbf{Left,} On the central-bank monetary-policy task (Sec.~\ref{sec:expr-central-bank}), OptScientist attains higher held-out policy-stance accuracy with substantially lower output-token consumption.
    \textbf{Right,} On human decisions under risk and ambiguity (Sec.~\ref{sec:expr-human-choice}), OptScientist reaches competitive held-out predictive accuracy with substantially fewer output tokens.
    Across both domains, OptScientist achieves a favourable accuracy--token-efficiency trade-off.
    }
    \label{fig:llm_baseline}
\end{figure}

We finally compared the token efficiency of OptScientist with three LLM baselines on the two real-world discovery tasks in Secs.~\ref{sec:expr-human-choice} and~\ref{sec:expr-central-bank}. All methods received the same research description, variable information, and training data; for the LLM baselines, we evaluated four maximum output-token settings and recorded held-out accuracy and cumulative output-token consumption.

\textbf{Token efficiency.}
OptScientist achieves near-best held-out predictive performance with substantially fewer LLM output tokens (Figure~\ref{fig:llm_baseline}). On central-bank monetary-policy decisions, OptScientist achieves \(78.6\%\) held-out policy-stance accuracy using 5,976 output tokens, exceeding the strongest LLM-based baseline result of \(70.5\%\), while the baselines consume approximately 16,700--29,600 tokens. On human decisions under risk and ambiguity, OptScientist approaches the
apparent predictive ceiling of approximately 83\% held-out accuracy while using only 6,396 output tokens, compared with approximately 15,800--45,800 tokens for the LLM baselines.

\section{Data and Evaluation} \label{sec:data}

\subsection{The OptFeynman} \label{sec:optfeynman}

\textbf{Benchmark design.}
To evaluate whether OptScientist can recover symbolic optimization laws, we construct \emph{OptFeynman}, a benchmark derived from the AI Feynman equation collection \citep{udrescu2020ai}. The original benchmark asks whether a method can recover an input--output equation. OptFeynman poses a different question: whether a method can recover the symbolic optimization problem that generates observed outcomes. Each analytic equation is therefore treated as a latent objective landscape, and the observed data are sampled as noisy outcomes of this landscape.

\textbf{Optimization construction.}
For each multivariate Feynman equation, we separate its variables into a context \(x\)
and a decision variable \(y\). The original equation defines the
ground-truth objective,
\[
    U(x,y)=f(x,y).
\]
Single-variable equations are excluded because they do not define a context-dependent
decision problem, leaving 99 OptFeynman tasks. Context variables are sampled uniformly from
the ranges specified in the original Feynman benchmark. For each task, we generate a set of unique contexts and partition them at the
context level into training, validation and test sets. All stochastic outcomes
generated from a given context are assigned to the same split. Thus, performance reflects generalization across unseen \(x\)-contexts, with stochastic outcomes from each context kept entirely within one split.

\textbf{Gibbs sampling.}
Observed outcomes are generated from the optimization landscape through Gibbs sampling.
In the discrete version, the decision variable is first discretized into five candidate actions
within its original range,
\[
    y_k = y_{\min} + k\frac{y_{\max}-y_{\min}}{K-1},
    \qquad k=0,\ldots,K-1,\quad K=5.
\]
For each context \(x_i\), the ground-truth objective is evaluated on these candidates. The
resulting scores are scaled by the standard deviation of the training objective values,
estimated only from training contexts, and converted into a Gibbs choice distribution,
\[
    P(y=k\mid x_i)
    =
    \frac{\exp(U(x_i,y_k)/(cT))}
    {\sum_{\ell=0}^{K-1}\exp(U(x_i,y_\ell)/(cT))}.
\]
Observed labels are then sampled from this categorical distribution, with
multiple labels drawn for each context.

The continuous version follows the same principle, while keeping \(y\) continuous on its
original interval \([y_{\min},y_{\max}]\). For each context, the target conditional density is
defined by the exponentiated objective,
\[
    p(y\mid x_i)
    \propto
    \exp\left(U(x_i,y)/(cT)\right),
    \qquad y\in [y_{\min},y_{\max}].
\]
Because this density is not analytically tractable for arbitrary Feynman equations, it is approximated on a dense grid over the decision interval. Grid points are assigned normalized Gibbs weights, a grid point is sampled as the outcome, and the sampled value is optionally perturbed within its local grid cell and clipped back to the valid interval. The continuous setting uses the same context-level construction as the discrete setting, but draws a larger set of continuous outcomes from each context. Thus, the same symbolic objective is instantiated either over a finite action set
or over a continuous decision interval.

\textbf{Noisy-context version.}
We also construct a noisy-context version of OptFeynman. Starting from the clean dataset,
Gaussian noise is added only to the training contexts,
\[
    \widetilde{x}_{ij}=x_{ij}+\epsilon_{ij},
    \qquad
    \epsilon_{ij}\sim\mathcal{N}(0,\sigma_j^2).
\]
The default noise scale is one percent of the sampling range of each context variable. The
validation and test contexts remain clean, and the ground-truth objective, sampled outcomes,
true scores, and evaluation files are unchanged. This setting tests whether OptScientist can
recover a stable underlying law when the observations used for learning are corrupted by
measurement noise.

\textbf{Evaluation.}
Evaluation is based on the utility-representation view of optimization. By Debreu's
continuous utility representation theorem, preferences can be represented by a utility
function when the preference relation is complete, transitive, and continuous
\citep{debreu1953representation}.

\begin{theorem}[Continuous utility representation, Debreu]
Let \(\mathcal{Z}\subseteq\mathbb{R}^n\), and let \(\succeq\) be a complete, transitive,
and continuous preference relation on \(\mathcal{Z}\). Then there exists a continuous
function \(u:\mathcal{Z}\rightarrow\mathbb{R}\) such that, for all
\(z_1,z_2\in\mathcal{Z}\),
\[
    z_1\succeq z_2
    \quad\Longleftrightarrow\quad
    u(z_1)\geq u(z_2).
\]
\end{theorem}

This result motivates our evaluation: an equivalent optimization problem should preserve the
preference structure induced by the latent objective, while also explaining its stochastic
realizations. Since OptFeynman provides the generating objective
\(U(x,y)\), we evaluate whether the learned score and the distilled symbolic score recover
the same ordinal and probabilistic structure.

For each context \(x_i\), let \(\mathcal{Y}_i=\{y_{i1},\ldots,y_{iK}\}\) be the candidate
outcome set. In discrete tasks, \(\mathcal{Y}_i\) contains the five candidate actions; in
continuous tasks, it contains sampled continuous candidates. The ground-truth objective
induces an ordered list \(\pi_i=(j_1,\ldots,j_K)\),
\[
    U(x_i,y_{ij_1}) \geq U(x_i,y_{ij_2}) \geq \cdots \geq U(x_i,y_{ij_K}),
\]
with \(\widehat{\pi}_i\) defined analogously by the recovered score \(\widehat{S}\).
Exact preference-order accuracy measures whether \(\widehat{\pi}_i=\pi_i\), and pairwise
accuracy measures whether each pairwise comparison has the same direction under \(U\) and
\(\widehat{S}\). These metrics test whether the recovered score represents the same
optimization problem up to order-preserving transformations.

Because the observations are generated by Gibbs sampling, we also evaluate the induced
stochastic landscape. For discrete tasks, the Gibbs distribution is computed over the five
candidate actions. For continuous tasks, it is approximated on a dense grid over the decision
interval. We compare the true and predicted distributions using KL divergence:
\[
    D_\mathrm{KL}(P\Vert \widehat{P})
    =
    \frac{1}{N}\sum_i\sum_{y\in\mathcal{Y}_i}
    P(y\mid x_i)\log\frac{P(y\mid x_i)}{\widehat{P}(y\mid x_i)} .
\]
Together, preference recovery evaluates whether the discovered expression captures the
equivalent optimization law, while distribution recovery evaluates whether it also captures
the stochastic strength of the generating landscape.

\subsection{Real-world Datasets} \label{sec:real-world-datasets}

Synthetic benchmarks test recovery when the generating optimization law is known.
Real-world data test discovery under noise, redundancy and incomplete theoretical
specification. We evaluate OptScientist on two domains: human risky choice and
central-bank monetary decisions.

\textbf{Human decisions under risk and ambiguity.}
The first dataset is the Choice Prediction Competition 2018 (CPC18), a large-scale
dataset of repeated human choices between monetary prospects under risk and ambiguity
\citep{plonsky2019cpc18}. In each problem, a participant repeatedly chooses between
two options, \(A\) and \(B\). The target variable is the observed binary choice, indicating
whether option \(B\) is selected.
CPC18 contains several groups of variables. The main lottery variables describe the
objective choice environment, including outcomes, probabilities, ambiguity and correlation
between the two options. Identifier variables, such as participant and game identifiers, organize repeated decisions
across participants and problems. Framing variables describe how the choice problem and
feedback are presented. Demographic variables provide participant-level covariates. Because
each participant faces repeated decisions with feedback, we also construct
history variables indicating whether a previous round exists and which option was
chosen in the previous round.
This dataset tests whether OptScientist can recover the optimization law behind human decisions under risk, ambiguity and sequential feedback.

\textbf{Central-bank monetary decisions.}
The second dataset studies central-bank monetary decisions from quarterly
macroeconomic and financial conditions. The outcome \(y\) is a five-class
quarterly policy stance, ranging from large easing to large tightening. We
construct \(y\) from the quarter-to-quarter change in the average federal funds
target rate. Before 2008, the daily target rate is taken from the single federal
funds target-rate series. After 2008, it is computed as the midpoint of the
target-rate corridor. The daily target rate is averaged within each quarter, and
the quarterly change is measured in basis points. A decrease of at least 50 basis
points is coded as \(-2\), a decrease between 25 and 50 basis points as \(-1\),
a change within 25 basis points as \(0\), an increase between 25 and 50 basis
points as \(1\), and an increase of at least 50 basis points as \(2\).

The context \(x\) contains twelve quarterly macro-financial variables. These
include PCE inflation, core PCE inflation, unemployment and real GDP growth;
two policy-inertia variables, given by the previous average target rate and the
previous quarterly target-rate change; three financial-market variables, given
by the Baa--Treasury credit spread, the 10-year--3-month term spread and the
CBOE Volatility Index (VIX);
and three financial-stress measures, given by the St. Louis Fed Financial Stress
Index, the Chicago Fed National Financial Conditions Index and the Adjusted
National Financial Conditions Index. The period variable is retained for temporal
indexing and data splitting.
This dataset tests whether OptScientist can recover a compact symbolic
optimization law behind macroeconomic policy decisions.

All policy-rate, macroeconomic and financial variables are obtained from
Federal Reserve Economic Data (FRED), maintained by the Federal Reserve Bank
of St. Louis. The federal funds target-rate series and target-range bounds are
used to construct the policy outcome, while the macroeconomic, financial-market
and financial-stress variables are taken from their corresponding FRED series.

\section{Conclusion and Limitations} \label{sec:discussion}

We introduced OptScientist, a framework for automatically discovering compact
symbolic optimization laws from observational data. OptScientist combines a
model that learns and distills latent optimization structure with an agentic
scientific-discovery system that incorporates theoretical priors, searches for
candidate laws, evaluates empirical evidence, and abstracts supported
mechanisms. On OptFeynman, OptScientist recovers ground-truth optimization laws
across a broad range of physics-derived problems, substantially outperforms all evaluated baselines, and remains robust to noisy
contextual information. Applied to human risky choice and central-bank monetary
policy, it discovers interpretable optimization laws with competitive or
stronger held-out predictive performance and compact symbolic representations.
On the two real-world tasks, OptScientist also uses substantially fewer output tokens than LLM-based automated model-discovery approaches. These results establish symbolic
optimization-law discovery as a new direction for automated scientific
discovery and provide a general framework for learning compact optimization
principles underlying observed phenomena.

\textbf{Limitations and Future Research.} When the structure of ground-truth models deviates substantially from an optimization formulation, our approach is likely to fail. Although,
mathematically, a broad class of laws not originally formulated as optimization
problems can be represented in optimization form, the OptFeynman results show
that OptScientist can recover the original laws from such
representations. Future research is needed to characterize the limits of the optimization assumption and identify the classes of models that remain discoverable within a feasible computational budget.

\newpage





\bibliography{references}
\bibliographystyle{open_preprint_iclr}

\newpage

\appendix

\startcontents[appendix]

\section*{Appendix Contents}
\printcontents[appendix]{}{1}{}

\newpage

\section{Additional Architecture Details}
\label{sec:add-architecture}

\subsection{OptScientist Model}
\label{sec:add-optscientist-model}

The OptScientist Model implements the two-stage discovery procedure introduced
in Sec.~\ref{sec:optscientist-model}. The Optimization Learner
(\emph{OptLearner}) first learns to represent observational data as an optimization
problem, and the Optimization Discoverer (\emph{OptDiscover}) subsequently
extracts its symbolic form. The discrete- and continuous-outcome implementations
share the same model architecture, optimization procedure, validation protocol,
and symbolic-distillation pipeline. They differ primarily in how candidate
decisions are constructed for the OptLearner training objective.

\begin{algorithm}[htbp]
\caption{OptScientist Model}
\label{alg:add-optscientist-model}
\begin{algorithmic}[1]
\Require Training data $\mathcal{D}_{\mathrm{train}}$,
validation data $\mathcal{D}_{\mathrm{val}}$,
decision domain $\mathcal{Y}$,
outcome type $r\in\{\mathrm{discrete},\mathrm{continuous}\}$,
OptLearner configuration $\mathcal{M}$,
symbolic search spaces $\Omega$
\Ensure Symbolic optimization law $\widehat{\mathcal{L}}$

\State Initialize OptLearner $S_{\theta}(x,y)$ from $\mathcal{M}$
\State Initialize best validation loss $\ell^{*}\gets\infty$

\For{$e=1,\ldots,E_{\max}$}
    \For{minibatch $\mathcal{B}\subset\mathcal{D}_{\mathrm{train}}$}
        \For{each $(x_i,y_i)\in\mathcal{B}$}
            \If{$r=\mathrm{discrete}$}
                \State $\mathcal{A}_i\gets\mathcal{Y}$
                \Comment{enumerate all feasible decisions}
            \Else
                \State Sample
                $\widetilde y_{i1},\ldots,\widetilde y_{iK}
                \sim\mathrm{Unif}(\mathcal{Y})$
                \State $\mathcal{A}_i
                \gets
                \{y_i,\widetilde y_{i1},\ldots,\widetilde y_{iK}\}$
            \EndIf
        \EndFor

        \State Compute
        \[
        \mathcal{L}
        =
        -\frac{1}{|\mathcal{B}|}
        \sum_{i\in\mathcal{B}}
        \log
        \frac{
        \exp(S_{\theta}(x_i,y_i)/\tau)
        }{
        \sum_{y\in\mathcal{A}_i}
        \exp(S_{\theta}(x_i,y)/\tau)
        }
        \]
        \State Update $\theta$ using AdamW
    \EndFor

    \State Evaluate validation loss $\ell_{\mathrm{val}}$
    \If{$\ell_{\mathrm{val}}<\ell^{*}$}
        \State Save OptLearner checkpoint
        \State $\ell^{*}\gets\ell_{\mathrm{val}}$
        \State Reset early-stopping counter
    \Else
        \State Increment early-stopping counter
    \EndIf
    \If{early-stopping patience is reached}
        \State \textbf{break}
    \EndIf
\EndFor

\State Restore the checkpoint with lowest validation loss
\State Freeze OptLearner
\State Construct symbolic-distillation samples
$\mathcal{D}_{\mathrm{distill}}$

\For{each active component
$H_{\theta,j}\in\{U_{\theta},C_{\theta},T_{\theta}\}$}
    \State Evaluate $H_{\theta,j}$ on
    $\mathcal{D}_{\mathrm{distill}}$
    \State Search $\mathcal{F}_{\Omega_j}$ for symbolic
    approximations of $H_{\theta,j}$
    \State Retain the simplest expression with near-minimal
    distillation loss
\EndFor

\State Remove additive $x$-only terms from the distilled objective
\State Assemble the retained components according to the learned
optimization structure
\State \Return $\widehat{\mathcal{L}}$
\end{algorithmic}
\end{algorithm}

\paragraph{OptLearner architecture.}
OptLearner uses the Behavior Learning Network (BL) as its backbone to learn
a scalar optimization landscape over context--outcome pairs. For each context
$x$ and candidate outcome $y$, the network returns a score
$S_{\theta}(x,y)$ that measures how strongly $y$ is preferred under the
latent optimization problem.

Building on the BL backbone, OptLearner represents observational data as a
single optimization problem. Its early layers denoise and compress observations,
capturing interactions and latent features that may be difficult to express
in the original coordinates. The final layer consists of a single BL block, forming an optimization representation bottleneck with learnable objective and optional constraint components, each potentially comprising multiple branches,
\begin{equation}
\max_{y\in\mathcal{Y}}
\underbrace{U_{\theta}(x,y)}_{[\lambda_0]}
\quad
\mathrm{s.t.}\quad
\underbrace{C_{\theta}(x,y)}_{[\lambda_1]}\leq 0,
\qquad
\underbrace{T_{\theta}(x,y)}_{[\lambda_2]}=0.
\label{eq:add-optscientist-final-block}
\end{equation}
Here $\lambda_0$ controls the scale of the objective, while $\lambda_1$ and
$\lambda_2$ control the strengths of the inequality and equality terms.
This bottleneck provides a structured optimization representation for
subsequent symbolic discovery. When the input variables are already
meaningful, OptLearner primarily denoises the optimization landscape.
For high-dimensional or weakly structured inputs, it additionally extracts
and compresses relevant features before forming the learned optimization
representation.

In the default implementation, OptLearner contains two BL layers with widths
$[128,1]$. The first layer uses one $U$ branch and one $C$ branch, and the
default final block contains a single $U$ branch. This configuration provides
the default single-objective form used in the main OptFeynman experiments.
Task-specific configurations can additionally activate $C$ and $T$ in the
final block, as in the constrained OptScientist variants used in our
experiments.

\paragraph{Unified learning objective.}
The discrete and continuous implementations can be written using the same
candidate-ranking objective. For observation $(x_i,y_i)$, let
$\mathcal{A}_i$ denote the set of candidate decisions considered during that
training step. OptLearner minimizes
\begin{equation}
\mathcal{L}_{\mathrm{OptLearner}}
=
-\frac{1}{N}
\sum_{i=1}^{N}
\log
\frac{
\exp\!\left(S_{\theta}(x_i,y_i)/\tau\right)
}{
\sum_{y\in\mathcal{A}_i}
\exp\!\left(S_{\theta}(x_i,y)/\tau\right)
}.
\label{eq:add-optscientist-unified-loss}
\end{equation}
For discrete outcomes,
$\mathcal{A}_i=\mathcal{Y}$ and the model evaluates the complete finite
decision set, giving the cross-entropy objective in
Eq.~\eqref{eq:add-optscientist-unified-loss}. For continuous outcomes, OptLearner constructs
a sampled candidate set,
\begin{equation}
\mathcal{A}_i
=
\left\{
y_i,
\widetilde y_{i1},\ldots,\widetilde y_{iK}
\right\},
\qquad
\widetilde y_{ik}
\sim
\mathrm{Unif}(\mathcal{Y}),
\end{equation}
where the observed outcome is compared with $K$ alternatives sampled
uniformly from the feasible decision range. This gives the sampled
contrastive Gibbs objective
in Eq.~\eqref{eq:continuous-loss}. Both settings
therefore train the same latent optimization score by assigning higher scores
to observed outcomes than to feasible alternatives under the same context.


OptLearner is optimized on the training split and selected using validation
loss. The best checkpoint is updated whenever validation loss improves.
Training terminates after the specified number of consecutive epochs without
validation improvement or when the maximum number of epochs is reached. The
checkpoint with the lowest validation loss is then restored and frozen before
symbolic discovery.

\paragraph{OptDiscover implementation.}
After OptLearner is trained, its parameters are frozen. OptDiscover extracts
the symbolic form of the learned optimization problem by distilling the
components of the optimization representation bottleneck. For each active
component
\[
H_{\theta,j}\in
\{U_{\theta},C_{\theta},T_{\theta}\},
\]
the model generates a distillation dataset over context--decision pairs and
records the corresponding learned component values. Symbolic regression then
searches for an expression
\begin{equation}
\widehat H_j
=
\arg\min_{F\in\mathcal{F}_{\Omega_j}}
\frac{1}{M}
\sum_{m=1}^{M}
\left[
F(x_m,y_m)-H_{\theta,j}(x_m,y_m)
\right]^2,
\label{eq:add-optscientist-distillation}
\end{equation}
where $\mathcal{F}_{\Omega_j}$ denotes the symbolic search space for the
$j$-th component.

The symbolic language $\Omega_j$ is task-dependent. We use a compact base
operator set by default, and additional binary or unary operators can be
introduced when required by the scientific problem or by explicit search
priors. This design allows the symbolic hypothesis space to reflect the
mathematical structure of each discovery task.

Symbolic search is performed separately for the active components of the
optimization representation bottleneck. The resulting expressions are
assembled according to the learned optimization structure, yielding a
compact symbolic optimization problem with an explicit objective and
optional constraints. Among expressions with near-minimal distillation
loss, OptDiscover selects the expression with the lowest effective symbolic
complexity. For the objective component, additive terms that depend only
on $x$ are subsequently removed because they remain constant across
candidate decisions and therefore do not affect the optimizer. The
retained symbolic components define the final discovered optimization law.

Algorithm~\ref{alg:add-optscientist-model} summarizes the common
implementation for discrete and continuous outcomes.

\paragraph{Implementation and hyperparameters.}
Table~\ref{tab:add-optscientist-hyperparameters} reports the default
implementation settings used in our experiments. These values specify the
base OptLearner and OptDiscover configuration. The active $U/C/T$ structure
and symbolic operator language can vary across task-specific configurations.
The operator entries in the table report the default symbolic search language.

\begin{table}[t]
\centering
\scriptsize
\setlength{\tabcolsep}{4.5pt}
\renewcommand{\arraystretch}{1.18}
\caption{
\textbf{Default OptScientist Model hyperparameters.}
The discrete and continuous implementations use the same OptLearner
architecture and training procedure, with different candidate construction
for the learning objective. Symbolic operators are task-dependent; the table
reports the default base language, and additional binary or unary operators
can be introduced when required by the discovery problem. The default final
block contains a single objective component $U$, while task-specific
configurations can additionally activate $C$ and $T$ components.
}
\label{tab:add-optscientist-hyperparameters}
\begin{tabular}{lcc}
\toprule
\textbf{Hyperparameter} &
\textbf{Discrete} &
\textbf{Continuous} \\
\midrule

\multicolumn{3}{l}{\textit{OptLearner architecture}} \\
Number of BL layers
    & 2 & 2 \\
BL layer widths
    & $[128,1]$ & $[128,1]$ \\
Hidden-layer $(U,C,T)$ branches
    & $(1,1,0)$ & $(1,1,0)$ \\
Default final-layer $(U,C,T)$ branches
    & $(1,0,0)$ & $(1,0,0)$ \\
$U$ activation
    & Identity & Identity \\
$C$ activation
    & ReLU & ReLU \\
Output-head bias
    & Yes & Yes \\
Decision encoding
    & Scalar & Scalar \\

\addlinespace
\multicolumn{3}{l}{\textit{OptLearner training}} \\
Learning objective
    & Cross-entropy
    & Sampled contrastive Gibbs \\
Contrastive alternatives $K$
    & -- & 32 \\
Temperature $\tau$
    & 1.0 & 1.0 \\
Alternative-decision sampling
    & -- & Uniform over feasible range \\
Optimizer
    & AdamW & AdamW \\
Learning rate
    & $2\times10^{-3}$ & $2\times10^{-3}$ \\
Weight decay
    & $10^{-5}$ & $10^{-5}$ \\
Batch size
    & 128 & 128 \\
Maximum epochs
    & 480 & 480 \\
Early-stopping patience
    & 25 & 25 \\
Random seed
    & 42 & 42 \\
Model-selection criterion
    & Validation loss & Validation loss \\

\addlinespace
\multicolumn{3}{l}{\textit{OptDiscover symbolic distillation}} \\
Active final components
    & Task-dependent; default $U$
    & Task-dependent; default $U$ \\
Distillation target
    & Activated component output
    & Activated component output \\
Binary operators
    & Task-dependent; default $\{+,-,\times\}$
    & Task-dependent; default $\{+,-,\times\}$ \\
Unary operators
    & Task-dependent
    & Task-dependent \\
Symbolic-search iterations
    & 200 & 200 \\
Maximum expression size
    & 24 & 24 \\
Distillation samples
    & 1,000 & 800 \\
Simplicity tolerance
    & 0.042 & 0.05 \\
Model-selection rule
    & Best & Best \\
\bottomrule
\end{tabular}
\end{table}

The symbolic-search settings provide a base configuration that can be adapted
to the scientific structure of each task. In particular, $\Omega$ can be
expanded with task-relevant operators before OptDiscover is executed. This
allows the OptLearner architecture to remain common across tasks while the
symbolic hypothesis space adapts to each scientific domain.

\begin{figure}[htbp]
    \centering
    \includegraphics[width=0.92\linewidth]{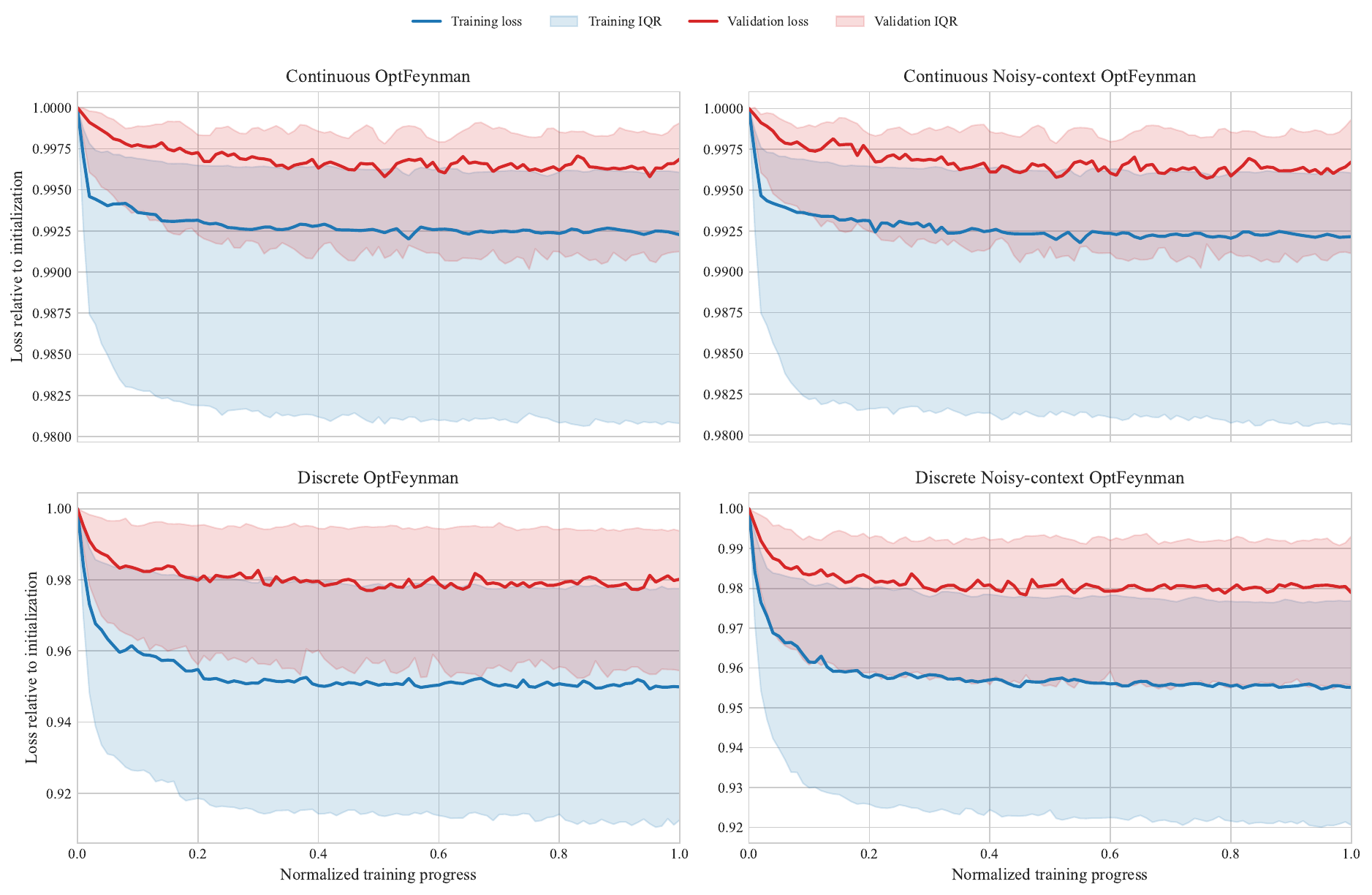}
    \caption{
    \textbf{OptLearner learning curves on OptFeynman.}
    Training and validation losses for continuous and discrete OptFeynman tasks,
    with and without noisy context variables. Loss is the OptLearner objective
    defined in the main text: \(L_{\mathrm{cont}}\) for continuous outcomes and
    \(L_{\mathrm{disc}}\) for discrete outcomes. For each task, losses are
    normalized by their values at the first epoch, and training progress is
    normalized from the first epoch to the epoch selected by validation-based
    early stopping. Curves show the median across tasks and shaded regions show
    the interquartile range.
    }
    \label{fig:add-optscientist-learning-curves}
\end{figure}

\subsection{OptScientist Agent}
\label{sec:add-optscientist-agent}

The OptScientist Agent implements the iterative scientific-discovery process introduced in Sec.~\ref{sec:optscientist-agent}. A Research Orchestrator maintains the research state and coordinates three specialized agents: a Theory Prior Agent, an Experiment Agent, and a Scientific Critic and Theory Abstraction Agent. Together, these components connect scientific reasoning with model training, symbolic discovery and empirical evaluation, allowing candidate optimization laws to be progressively proposed, tested, adjudicated and abstracted.

The architecture is organized around a shared research state that accumulates information across discovery rounds. This state contains the research question, variable definitions, theoretical hypotheses, fixed data split, search configurations, discovered symbolic laws, evaluation results and unresolved scientific questions. Maintaining these elements in a common state allows later stages to trace each candidate law back to the assumptions, variables and empirical evidence that produced it. It also enables the system to preserve informative failures: unstable symbolic structures, weakly supported mechanisms and missing explanatory variables can directly shape the next round of discovery.

Each specialized agent operates on a role-specific view of this state and returns structured outputs to the Research Orchestrator. Scientific priors therefore enter the workflow as explicit hypotheses and search constraints; empirical results enter through executed OptScientist runs; and scientific interpretations are formed from the accumulated evidence across candidate laws.

\textbf{Embedding the OptScientist Model in the discovery loop.}
The OptScientist Model forms the empirical core of the agentic discovery system. In each discovery round, the Theory Prior Agent converts scientific knowledge into a structured set of candidate hypotheses, variables and search priors. These outputs can include newly constructed variables, transformations of observed variables, candidate mechanisms, preferred symbolic operators and restrictions on admissible functional forms. The Experiment Agent combines these proposals with model and training parameters to construct a bounded set of executable OptScientist runs.

Concretely, each run is represented by a configuration
\[
Q_k =
\left(
\mathcal{V}_k,\,
\mathcal{M}_k,\,
\Theta_k,\,
\Omega_k
\right),
\]
where $\mathcal{V}_k$ denotes the observed and theory-derived variables supplied to the model, $\mathcal{M}_k$ specifies the OptScientist representation to be fitted, $\Theta_k$ contains the corresponding model, training and search parameters, and $\Omega_k$ defines the symbolic language used for distillation. For example, a prior analysis of the data informed by the scientific literature may suggest an expected-reward variable or a particular interaction between observed quantities. The Experiment Agent adds these variables to the corresponding model input, assigns an OptLearner architecture and training configuration, specifies the operators available to OptDiscover, and submits the resulting configuration as an OptScientist run. Different hypotheses, variable sets or parameter settings can therefore generate complementary runs within the same discovery round.

\begin{figure}[t]
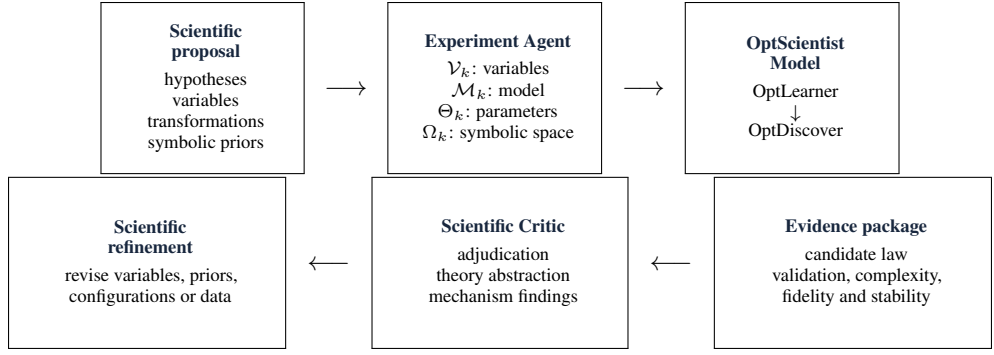

\centering
\setlength{\fboxsep}{3pt}
\fbox{
\parbox[c][5.8em][c]{0.17\linewidth}{
\centering
\scriptsize
\textbf{Scientific}\\[-0.05em]
\textbf{proposal}\\[0.35em]
hypotheses\\
variables\\
transformations\\
symbolic priors
}}
\hspace{0.25em}
$\longrightarrow$
\hspace{0.25em}
\fbox{
\parbox[c][5.8em][c]{0.18\linewidth}{
\centering
\scriptsize
\textbf{Experiment Agent}\\[0.35em]
$\mathcal{V}_k$: variables\\
$\mathcal{M}_k$: model\\
$\Theta_k$: parameters\\
$\Omega_k$: symbolic space
}}
\hspace{0.25em}
$\longrightarrow$
\hspace{0.25em}
\fbox{
\parbox[c][5.8em][c]{0.18\linewidth}{
\centering
\scriptsize
\textbf{OptScientist}\\[-0.05em]
\textbf{Model}\\[0.35em]
OptLearner\\
$\downarrow$\\[-0.15em]
OptDiscover
}}
\vspace{0.8em}
\fbox{
\parbox[c][5.8em][c]{0.24\linewidth}{
\centering
\scriptsize
\textbf{Scientific}\\[-0.05em]
\textbf{refinement}\\[0.35em]
revise variables, priors,\\
configurations or data
}}
\hspace{0.3em}
$\longleftarrow$
\hspace{0.3em}
\fbox{
\parbox[c][5.8em][c]{0.22\linewidth}{
\centering
\scriptsize
\textbf{Scientific Critic}\\[0.35em]
adjudication\\
theory abstraction\\
mechanism findings
}}
\hspace{0.3em}
$\longleftarrow$
\hspace{0.3em}
\fbox{
\parbox[c][5.8em][c]{0.24\linewidth}{
\centering
\scriptsize
\textbf{Evidence package}\\[0.35em]
candidate law\\
validation, complexity,\\
fidelity and stability
}}
\caption{
\textbf{Embedding the OptScientist Model in the agentic discovery loop.}
The Theory Prior Agent produces scientific hypotheses, candidate variables and symbolic priors. The Experiment Agent combines these proposals with model and search parameters into executable configurations $Q_k$. Each configuration is passed through OptLearner and OptDiscover to produce candidate symbolic optimization laws and associated empirical evidence. The Scientific Critic evaluates this evidence, abstracts the supported laws into a more general theoretical form and qualitative mechanisms, and returns refinements that can modify the variables, priors, model configurations or data used in the next discovery round.
}
\label{fig:optscientist-model-agent-interface}
\end{figure}

Figure~\ref{fig:optscientist-model-agent-interface} summarizes this interface. Scientific proposals are first converted into executable configurations. These configurations are passed to the OptScientist Model, which learns an optimization representation and distills it into symbolic laws. The resulting laws and their empirical evidence are then returned to the research state for scientific adjudication and refinement.

Each configuration is evaluated using the two-stage OptScientist Model introduced in Sec.~\ref{sec:optscientist-model}. Given the training data restricted to the variables in $\mathcal{V}_k$, OptLearner fits a scalar optimization landscape,
\[
S_{\theta_k}
=
\textsc{OptLearner}
\left(
\mathcal{D}_{\mathrm{train}}[\mathcal{V}_k],
\mathcal{M}_k,
\Theta_k
\right).
\]
The learned hierarchy can denoise observations, compress redundant information, or construct intermediate representations, which are then expressed as a single optimization problem through the optimization representation bottleneck. The fitted model therefore provides a data-driven representation of how candidate outcomes are ranked under the observed contexts.

After training, the OptLearner parameters are frozen. OptDiscover receives the learned optimization representation together with the symbolic specification $\Omega_k$ and distills its components into compact symbolic expressions,
\[
\mathcal{L}_k
=
\textsc{OptDiscover}
\left(
S_{\theta_k},
\Omega_k
\right).
\]
For a single-objective configuration, this procedure distills the learned objective directly. For a constrained configuration, the objective, inequality-constraint and equality-constraint components are distilled separately and then assembled according to the learned optimization structure. The symbolic-search specification controls the admissible operators and functional forms, while the distillation loss and complexity criterion determine which candidate expressions are retained.

The Experiment Agent collects the output of every successful run as an evidence package,
\[
\begin{aligned}
\mathcal{E}_k = \bigl(&
\text{validation performance},
\text{symbolic complexity},\\
&
\text{distillation fidelity},
\text{run stability},
\text{execution record}
\bigr).
\end{aligned}
\]
The pair $(\mathcal{L}_k,\mathcal{E}_k)$ is written back to the shared research state. Repeated symbolic searches can contribute several low-complexity expressions with near-optimal distillation loss, allowing alternative structures to remain available when the data do not identify a unique symbolic form. The resulting candidate set therefore records both the discovered mathematical expressions and the evidence supporting each expression.

The Scientific Critic and Theory Abstraction Agent receives these candidate laws together with their evidence records and the hypotheses that generated the corresponding runs. It compares candidates using predictive performance, symbolic complexity, fidelity to the learned optimization landscape and stability across repeated searches, and identifies the laws and mathematical structures that receive the strongest empirical support. It then abstracts these supported structures into a more general theoretical form and summarizes the corresponding qualitative mechanisms and scientific findings. The exact candidate expressions are retained alongside this abstraction, preserving the connection between the empirical laws and the higher-level theory derived from them.

Remaining uncertainty is translated into concrete modifications of the next discovery round. These may include introducing a new theoretically motivated variable, removing a redundant representation, changing the symbolic operator set, evaluating an alternative OptScientist configuration or requesting additional observations. The resulting theory abstraction, mechanism findings and refinement proposals are written back to the shared research state and used to define the subsequent round of discovery.

This interface creates a direct path from scientific reasoning to model execution and back to scientific interpretation. Theory-guided variables, hypotheses and search priors are compiled by the Experiment Agent into explicit OptScientist configurations; OptLearner and OptDiscover recover candidate optimization laws from the data; and the resulting evidence is evaluated by the Scientific Critic and Theory Abstraction Agent. The supported laws are then adjudicated, abstracted into a more general theoretical form and summarized as qualitative mechanisms and scientific findings. These conclusions, together with the remaining uncertainties, determine how the subsequent discovery round is refined.

Algorithm~\ref{alg:optscientist-agent} makes this interface explicit.

\begin{algorithm}[t]
\caption{OptScientist Agent}
\label{alg:optscientist-agent}
\begin{algorithmic}[1]
\Require Observational data $\mathcal{D}$, research question $q$, decision variable $y$, search budget $\mathcal{B}$
\Ensure Supported symbolic optimization laws $\mathcal{L}^{*}$ and scientific abstraction with qualitative mechanisms $\mathcal{A}^{*}$

\State Initialize shared research state $\mathcal{S}$
\State Fix training and validation splits
$\mathcal{D}_{\mathrm{train}},\mathcal{D}_{\mathrm{val}}$

\While{discovery continues}

    \State $(\mathcal{H},\mathcal{V},\Omega)
    \gets
    \textsc{TheoryPriorAgent}(\mathcal{S})$
    \Comment{hypotheses, variables, and search priors}

    \State Validate proposed variables $\mathcal{V}$
    \State Update $\mathcal{S}$ with accepted hypotheses and variables

    \State $\mathcal{Q}
    \gets
    \textsc{ExperimentAgent}
    (\mathcal{S},\mathcal{H},\mathcal{V},\Omega,\mathcal{B})$
    \Comment{executable OptScientist configurations}

    \State Initialize candidate-evidence set $\mathcal{C}\gets\varnothing$

    \For{each configuration $Q\in\mathcal{Q}$}

        \State $S_{\theta_Q}
        \gets
        \textsc{OptLearner}
        (\mathcal{D}_{\mathrm{train}},Q)$
        \Comment{learn latent optimization landscape}

        \State Freeze $S_{\theta_Q}$

        \State $\mathcal{L}_{Q}
        \gets
        \textsc{OptDiscover}
        (S_{\theta_Q},Q)$
        \Comment{distill candidate symbolic laws}

        \State $\mathcal{E}_{Q}
        \gets
        \textsc{Evaluate}
        (\mathcal{L}_{Q},S_{\theta_Q},\mathcal{D}_{\mathrm{val}})$
        \Comment{predictive, structural, and symbolic evidence}

        \State $\mathcal{C}
        \gets
        \mathcal{C}\cup
        \{(\mathcal{L}_{Q},\mathcal{E}_{Q},Q)\}$

    \EndFor

    \State Record candidate laws, evidence, and execution traces in $\mathcal{S}$

    \State $(\mathcal{L}^{*},\mathcal{A}^{*},\mathcal{R})
    \gets
    \textsc{ScientificCriticAndTheoryAbstraction}
    (\mathcal{S},\mathcal{H},\mathcal{C})$
    \Comment{adjudication, theory abstraction, mechanisms, and refinement}

    \State Update $\mathcal{S}$ with supported laws $\mathcal{L}^{*}$,
    theory abstraction and mechanism findings $\mathcal{A}^{*}$,
    associated evidence, and refinements $\mathcal{R}$

    \If{additional hypotheses, variables, or data are required}
        \State Initialize the next discovery round from $\mathcal{S}$
    \Else
        \State \textbf{break}
    \EndIf

\EndWhile

\State \Return $\mathcal{L}^{*},\mathcal{A}^{*}$
\end{algorithmic}
\end{algorithm}

\textbf{Research Orchestrator.}
The Research Orchestrator provides the control layer of the discovery process. It maintains the shared research state, coordinates information transfer among agents, manages the search budget and determines whether another discovery round is needed. The state records theoretical hypotheses, accepted variables, model configurations,
discovered symbolic expressions, their empirical evaluations, theory abstractions and
qualitative findings, creating a persistent scientific record across successive rounds.

This persistent state also provides continuity between scientific reasoning and empirical search. A hypothesis proposed early in the workflow remains associated with the variables introduced to test it, the model configurations used during discovery and the evidence obtained from the resulting symbolic laws. Similarly, unresolved mechanisms identified by the Scientific Critic can be translated into explicit requests for additional variables, alternative transformations or further model searches. The Orchestrator therefore allows the discovery process to accumulate evidence over time while preserving the provenance of intermediate scientific decisions.

Importantly, the Orchestrator also preserves the boundary between scientific reasoning and model execution. Agent outputs that define hypotheses, variables or search priors are converted into explicit OptScientist configurations before model fitting. Conversely, outputs from OptLearner and OptDiscover enter subsequent agent reasoning only through recorded candidate laws and their associated empirical evidence. This separation makes it possible to distinguish structures proposed as theoretical priors from structures that are subsequently supported by model fitting and symbolic discovery.

\textbf{Agent instructions.}
Each specialized agent is assigned a defined scientific role, information scope and output structure. The prompts emphasize testable hypotheses, executable empirical search and evidence-grounded scientific abstraction. Their principal instructions are summarized below.

\begin{center}
\fcolorbox{black!15}{black!3}{
\begin{minipage}{0.94\linewidth}
\small
\textbf{Theory Prior Agent}

\vspace{0.4em}
\textbf{Role.}
Act as a scientific theorist preparing an optimization-law discovery problem.

\textbf{Input.}
Research question, decision variable, available variables, compact training-side data structure and allowed variable transformations.

\textbf{Instructions.}
Identify established scientific theories and mechanisms relevant to the research question and connect them to observed variables without treating theoretical priors as empirical conclusions. Formulate a small set of explicit and testable mechanism hypotheses. Identify scientifically meaningful, target-independent variables or derived variables that provide observable representations of these mechanisms, and propose sparse symbolic-search priors over operators and functional structures when scientifically justified. Exclude variable constructions that depend on the observed target or validation and test information, and examine identifiability, redundancy and other theoretical risks. Express each proposal in a form that can be evaluated through subsequent model fitting and symbolic discovery.

\textbf{Output.}
Background theory, mechanism hypotheses, candidate variables, theoretical rationale, symbolic-search priors, potential risks and unresolved scientific questions.
\end{minipage}}
\end{center}

The Theory Prior Agent provides the scientific entry point to the discovery loop. Its role is particularly important when raw observations admit many possible representations. Domain concepts can suggest target-independent transformations that expose latent mechanisms, such as temporal differences, relative quantities, interaction terms or state variables. These proposals constrain the effective hypothesis space presented to symbolic discovery and provide an interpretable correspondence between discovered mathematical terms and candidate scientific mechanisms. Because the proposals remain explicit in the shared state, later stages can evaluate whether the resulting evidence supports, weakens or leaves each hypothesis unresolved.

The Theory Prior Agent does not determine the functional form of the final law. Its proposals define the scientific search space presented to OptScientist: which observed or derived variables may represent a mechanism, and which symbolic structures are scientifically plausible enough to prioritize during discovery. Whether those variables enter the final optimization law, and in what mathematical form, is determined subsequently by OptLearner and OptDiscover from the observational data.

\begin{center}
\fcolorbox{black!15}{black!3}{
\begin{minipage}{0.94\linewidth}
\small
\textbf{Experiment Agent}

\vspace{0.4em}
\textbf{Role.}
Translate scientific priors into an empirical search for symbolic optimization laws.

\textbf{Input.}
Theory priors, validated variables, task type, available OptScientist configurations, symbolic operators, fixed validation protocol and compute budget.

\textbf{Instructions.}
Organize the theoretical mechanisms, proposed variables and symbolic-search priors into a bounded set of complementary OptScientist configurations. For each configuration, invoke OptLearner to fit the latent optimization landscape on the fixed training data and then invoke OptDiscover to distil the trained optimization representation into compact symbolic laws. Evaluate the resulting laws under the fixed validation protocol. Repeat symbolic discovery where useful for assessing structural stability, and preserve low-complexity, near-optimal complete expressions from each successful symbolic-search run rather than retaining only a single run-level expression. When variable construction fails, repair only the implementation while preserving its scientific meaning and target independence. Record the resulting candidate expressions, predictive performance, symbolic complexity, distillation evidence, repeated-run evidence and execution records. Preserve alternative high-performing structures for subsequent scientific adjudication rather than selecting the final scientific law.

\textbf{Output.}
OptScientist configurations, candidate symbolic optimization laws, predictive metrics, symbolic complexity, distillation and repeated-run evidence, execution records and runtime issues.
\end{minipage}}
\end{center}

The Experiment Agent provides the empirical interface between scientific reasoning and the OptScientist Model. Candidate hypotheses are not evaluated by asking the language model to directly write a final equation. Instead, the agent converts each hypothesis into an executable discovery configuration specifying the candidate variables, model structure, symbolic operators and relevant search settings. This produces a small set of scientifically differentiated model runs that can be compared under the same training and validation protocol.

Each run then follows the fixed internal logic of the OptScientist Model. OptLearner receives the observational variables selected for that configuration and learns a scalar optimization landscape $S_{\theta}(\mathbf{x},\mathbf{y})$. Depending on the structure of the observations, OptLearner may primarily
denoise meaningful scientific variables or additionally extract and compress
features from high-dimensional or weakly structured inputs. Its final layer
consists of a single BL block, forming an optimization representation
bottleneck that represents the data as a single optimization problem.
OptDiscover then extracts the symbolic form of the learned optimization
problem.

After OptLearner has been trained, its parameters are frozen and OptDiscover operates on the learned representation rather than directly fitting the observed target. For a constrained representation, the learned objective, inequality-constraint and equality-constraint components are distilled separately and then assembled into a symbolic optimization problem. For a single-objective representation, the corresponding learned objective is distilled directly. The symbolic operators and structural priors proposed earlier in the discovery round determine the candidate symbolic language, while the distillation objective and complexity control determine which expressions are retained. This division of labour is important: the agent specifies the scientific search problem, whereas OptScientist learns and distills the optimization structure supported by the data.

The resulting evidence extends beyond predictive performance. Each symbolic-search run can contribute several low-complexity expressions with near-optimal distillation loss, allowing alternative structures to remain available when the symbolic evidence does not identify a unique expression. Candidate laws are therefore compared using their validation performance, symbolic complexity, fidelity to the learned optimization representation and stability across repeated searches. Recording these dimensions provides the Scientific Critic with evidence about both empirical adequacy and structural reliability, while preserving multiple competitive expressions when the available data do not clearly distinguish among them.

This explicit model interface also separates model discovery from scientific adjudication. The Experiment Agent records successful and unsuccessful runs but does not choose which expression should be interpreted as the final scientific law. Candidate-law adjudication, theory abstraction and mechanism-level interpretation occur only after the candidate expressions and their associated evidence have been returned to the shared research state.

\begin{center}
\fcolorbox{black!15}{black!3}{
\begin{minipage}{0.94\linewidth}
\small
\textbf{Scientific Critic and Theory Abstraction Agent}

\vspace{0.4em}
\textbf{Role.}
Evaluate discovered optimization laws as candidate scientific explanations and abstract their best-supported scientific structure.

\textbf{Input.}
Theory hypotheses, candidate symbolic laws, validation evidence, symbolic complexity, distillation evidence, repeated-run stability and available robustness evidence.

\textbf{Instructions.}
Compare candidate laws using their empirical evidence and characterize the level of support for each proposed mechanism. Adjudicate among candidate laws by jointly considering predictive performance, symbolic complexity, structural stability and consistency with plausible theoretical mechanisms, while treating theory as a prior rather than a substitute for empirical support. Preserve the exact selected candidate expressions and identify structures and qualitative mechanisms that recur across supported laws. Synthesize the supported candidates into a general-to-specific theory abstraction, progressing from a general optimization form to supported functional structure, exact parameterized laws and qualitative findings. Assess explicit research hypotheses, identify competing explanations, limitations and remaining failure modes, and propose additional variables, observations or targeted analyses that could resolve the remaining scientific uncertainty.

\textbf{Output.}
Candidate-law adjudication, supported symbolic laws, general-to-specific theory abstraction, qualitative mechanisms, hypothesis assessment, limitations and targeted suggestions for subsequent discovery.
\end{minipage}}
\end{center}

The Scientific Critic connects symbolic recovery to theory formation. Symbolic regression can produce several expressions with similar predictive performance, including algebraically related forms and model-specific residual terms. The Critic evaluates these expressions collectively and identifies structures that remain stable across supported candidates. Recurring structures provide evidence for mechanism-level conclusions, whereas unstable or weakly identified terms are retained as sources of uncertainty.

The Critic receives the output of OptScientist as empirical evidence rather than as a finished theory. This distinction allows it to compare not only the symbolic expressions themselves but also how faithfully they reproduce the learned optimization landscape, how well they generalize under the fixed validation protocol, and whether similar structures reappear across repeated discovery runs. Scientific interpretation is therefore conditioned on the evidence produced by the model rather than on the plausibility of an expression in isolation.

Theory abstraction is performed hierarchically. At the highest level, the Critic identifies the general optimization principle expressed by the supported laws, such as reward maximization, target tracking or minimization of a deviation cost. It then constructs a parameter-tied general form where common parameter structure is empirically supported, followed by the specific functional structure associated with interpretable mechanisms. The exact parameterized candidate laws are retained at the next level, preserving their empirical expressions, and the final level summarizes the supported qualitative findings. This hierarchy allows OptScientist to synthesize candidate laws into a more general form and qualitative mechanisms while preserving the detailed mathematical structure from which those conclusions were derived.

The Critic also converts unresolved evidence into new scientific questions. A missing mechanism may motivate a new derived variable; competing symbolic structures may motivate additional observations capable of distinguishing their predictions; and systematic prediction failures may indicate omitted context or an insufficient symbolic language. These outputs are returned to the shared research state and provide structured inputs for a subsequent discovery round. In the next round, the Theory Prior and Experiment Agents translate these unresolved questions into revised variables, hypotheses or OptScientist configurations, so that scientific critique produces a concrete change in the subsequent model search rather than remaining a purely textual assessment.

Overall, the architecture organizes OptScientist into an iterative sequence of theory formation, model-based empirical discovery, evidence evaluation and theory abstraction. The Theory Prior Agent establishes candidate mechanisms, scientifically meaningful representations and symbolic-search priors. The Experiment Agent converts these proposals into executable configurations and invokes the OptScientist Model. Within each run, OptLearner learns to represent observational data as an optimization problem through an optimization representation bottleneck, and OptDiscover then distills the learned representation into a compact symbolic optimization law. The Scientific Critic then adjudicates among the resulting candidates and synthesizes their supported structure into more general optimization forms and qualitative mechanisms. The Research Orchestrator preserves the connection between these stages and carries unresolved scientific questions across iterations. This division of labour places a specialized symbolic-discovery model at the empirical centre of the agentic workflow, allowing language-model agents to guide and interpret discovery without substituting linguistic generation for model-based evidence.

\section{Additional Experimental Details}
\label{sec:additional-experimental-details}

\subsection{Baselines}
\label{sec:baselines}

We compare OptScientist with standard machine-learning models, symbolic regression, optimization-structured models, domain-specific models, and three LLM-based automated model-discovery methods. The general baselines include Multilayer Perceptron (MLP) \citep{rumelhart1986learning}, Logistic Regression (LR) \citep{cox1958regression}, Random Forest (RF) \citep{breiman2001random}, eXtreme Gradient Boosting (XGBoost) \citep{chen2016xgboost}, Gradient-Boosted Decision Trees (GBDT) \citep{friedman2001greedy}, Symbolic Regression (SR) \citep{cranmer2023interpretable}, Neural Energy-Based Models (Neural EBM) \citep{lecun2006tutorial}, and Behavior Learning (BL) \citep{ma2026behavior}. The LLM-based baselines are an LLM-only researcher, GeCCo \citep{rmus2026generating} as a code-assisted LLM approach, and Model Discovery Agent \citep[MDA;][]{murphy2026model} as a Bayesian experiment-design approach.

\textbf{OptFeynman.}
For OptFeynman, we compare against MLP, LR, RF, XGBoost, and GBDT as standard machine-learning baselines, SR as a direct symbolic-discovery baseline, Neural EBM and BL as optimization-structured baselines, and the three LLM-based baselines.

\textbf{Human decisions under risk and ambiguity.}
For CPC18, we compare against established behavioural models, including expected-value logit, expected-utility logit, prospect-theory logit with fixed or fitted parameters, and mean--variance logit. We additionally include GBDT, MLP, RF, and XGBoost as machine-learning baselines, together with the three LLM-based baselines.

\textbf{Central-bank monetary decisions.}
For monetary-policy decisions, we compare against MLP, RF, XGBoost, and GBDT as machine-learning baselines, Neural EBM as an optimization-structured baseline, SR as a direct symbolic-discovery baseline, and the three LLM-based baselines.

\subsection{OptFeynman Ground-Truth Recovery}

\textbf{Statistical analysis.}
We report additional statistics for the OptFeynman experiments in
Figures~\ref{fig:optfeynman-standard} and~\ref{fig:optfeynman-noisyx}. The analysis uses results from the OptLearner score and the final OptScientist symbolic law produced by OptDiscover for 99 OptFeynman problems under each combination of decision type (discrete or continuous) and training condition. Throughout this section, \emph{OptLearner score} refers to the learned optimization score before symbolic distillation, and \emph{OptScientist law} refers to the final symbolic optimization law produced by OptDiscover. We use \emph{Clean} for the original OptFeynman setting and \emph{Noisy-context} for the noisy-context version defined in Sec.~\ref{sec:optfeynman}, in which Gaussian noise is added only to the training contexts. Each problem is treated as one observation when computing statistics across the benchmark. One duplicated record in the continuous Clean setting contained identical metrics and was counted once.

Recovery is evaluated using exact-order accuracy, pairwise preference accuracy, and the Gibbs distribution induced by the recovered score. Exact-order accuracy measures agreement between the complete preference order produced by the recovered and ground-truth objectives. Pairwise accuracy measures agreement across individual pairs of candidate outcomes. For continuous tasks, preference recovery is evaluated at \(K=10\), \(50\), and \(100\) candidate outcomes, and distribution recovery is evaluated on a dense grid over the decision interval.

Unless stated otherwise, reported uncertainty is the standard error across the 99 OptFeynman problems,
\[
    \mathrm{SE}=\frac{s}{\sqrt{n}},
    \qquad n=99,
\]
where \(s\) is the sample standard deviation across problems. Error bars in Figures~\ref{fig:optfeynman-standard}
and~\ref{fig:optfeynman-noisyx} use mean \(\pm 1\) SE. Medians and interquartile ranges are also reported for metrics with uneven distributions across problems. Perfect exact-order recovery denotes a task with exact-order accuracy equal to one.

\textbf{Perfect exact-order recovery.}
Table~\ref{tab:optfeynman-perfect} summarizes perfect ground-truth recovery. For discrete outcomes, OptScientist recovers the complete ground-truth preference order on \(88/99\) Clean problems and \(92/99\) Noisy-context problems. For continuous outcomes, it achieves perfect recovery on \(78/99\) Clean problems and \(76/99\) Noisy-context problems at \(K=10\), \(50\), and \(100\). The stability of these counts as \(K\) increases shows that the recovered laws from OptScientist continue to preserve the full ordering over increasingly dense sets of candidate outcomes.

The learned OptLearner score has substantially fewer perfectly recovered problems. In the continuous Clean setting, the count decreases from \(30/99\) at \(K=10\) to \(14/99\) at \(K=100\). This difference is also visible in the discrete setting, where OptLearner achieves perfect recovery on \(31/99\) Clean problems and \(27/99\) Noisy-context problems.

\begin{table}[htbp]
\centering
\caption{\textbf{Perfect exact-order recovery on OptFeynman.}
Entries report the number and percentage of the 99 tasks with exact-order accuracy equal to one. OptScientist law denotes the final symbolic optimization law produced by OptDiscover.} \vspace{2mm}
\label{tab:optfeynman-perfect}
\small
\setlength{\tabcolsep}{4.5pt}
\renewcommand{\arraystretch}{1.12}
\begin{tabular}{@{}llccc@{}}
\toprule
Decision type & Representation & Candidate set & Clean & Noisy-context\\
\midrule
Discrete
& OptScientist law
& \(K=5\)
& \(88/99\;(88.89\%)\)
& \(92/99\;(92.93\%)\) \\
Discrete
& OptLearner score
& \(K=5\)
& \(31/99\;(31.31\%)\)
& \(27/99\;(27.27\%)\) \\
\midrule
Continuous
& OptScientist law
& \(K=10\)
& \(78/99\;(78.79\%)\)
& \(76/99\;(76.77\%)\) \\
Continuous
& OptLearner score
& \(K=10\)
& \(30/99\;(30.30\%)\)
& \(30/99\;(30.30\%)\) \\
Continuous
& OptScientist law
& \(K=50\)
& \(78/99\;(78.79\%)\)
& \(76/99\;(76.77\%)\) \\
Continuous
& OptLearner score
& \(K=50\)
& \(20/99\;(20.20\%)\)
& \(23/99\;(23.23\%)\) \\
Continuous
& OptScientist law
& \(K=100\)
& \(78/99\;(78.79\%)\)
& \(76/99\;(76.77\%)\) \\
Continuous
& OptLearner score
& \(K=100\)
& \(14/99\;(14.14\%)\)
& \(15/99\;(15.15\%)\) \\
\bottomrule
\end{tabular}
\end{table}

\textbf{Continuous outcomes.}
Table~\ref{tab:optfeynman-continuous} reports continuous preference recovery. Under the Clean condition, mean OptScientist exact-order accuracy is \(0.900\) at \(K=10\), \(0.871\) at \(K=50\), and \(0.866\) at \(K=100\). The corresponding values under Noisy-context are \(0.902\), \(0.874\), and \(0.869\).

\begin{table}[htbp]
\centering
\caption{\textbf{Preference recovery for continuous OptFeynman problems.}
Values are mean \(\pm\) SE across 99 OptFeynman problems. Median and interquartile range refer to exact-order accuracy. Higher values indicate better recovery.} \vspace{2mm}
\label{tab:optfeynman-continuous}
\footnotesize
\setlength{\tabcolsep}{3.2pt}
\renewcommand{\arraystretch}{1.12}
\begin{tabular}{@{}lclccc@{}}
\toprule
Condition & \(K\) & Representation
& Exact order
& Median [Q1,Q3]
& Pairwise \\
\midrule
Clean & 10 & OptScientist law
& \(0.900\pm0.027\)
& \(1.000\,[1.000,1.000]\)
& \(0.963\pm0.015\) \\
Clean & 10 & OptLearner score
& \(0.825\pm0.026\)
& \(0.940\,[0.800,1.000]\)
& \(0.960\pm0.007\) \\
Clean & 50 & OptScientist law
& \(0.871\pm0.032\)
& \(1.000\,[1.000,1.000]\)
& \(0.963\pm0.015\) \\
Clean & 50 & OptLearner score
& \(0.743\pm0.035\)
& \(0.900\,[0.695,0.990]\)
& \(0.960\pm0.008\) \\
Clean & 100 & OptScientist law
& \(0.866\pm0.033\)
& \(1.000\,[1.000,1.000]\)
& \(0.963\pm0.015\) \\
Clean & 100 & OptLearner score
& \(0.719\pm0.035\)
& \(0.880\,[0.665,0.970]\)
& \(0.960\pm0.008\) \\
\midrule
Noisy-context & 10 & OptScientist law
& \(0.902\pm0.026\)
& \(1.000\,[1.000,1.000]\)
& \(0.959\pm0.014\) \\
Noisy-context & 10 & OptLearner score
& \(0.826\pm0.026\)
& \(0.950\,[0.805,1.000]\)
& \(0.960\pm0.008\) \\
Noisy-context & 50 & OptScientist law
& \(0.874\pm0.030\)
& \(1.000\,[1.000,1.000]\)
& \(0.959\pm0.014\) \\
Noisy-context & 50 & OptLearner score
& \(0.743\pm0.035\)
& \(0.910\,[0.695,0.990]\)
& \(0.960\pm0.008\) \\
Noisy-context & 100 & OptScientist law
& \(0.869\pm0.031\)
& \(1.000\,[1.000,1.000]\)
& \(0.958\pm0.014\) \\
Noisy-context & 100 & OptLearner score
& \(0.722\pm0.035\)
& \(0.880\,[0.665,0.980]\)
& \(0.960\pm0.008\) \\
\bottomrule
\end{tabular}
\end{table}

\begin{table}[htbp]
\centering
\caption{\textbf{Distribution recovery for continuous OptFeynman problems.}
KL divergence is computed between the ground-truth and recovered Gibbs distributions on the dense test grid. Values are mean \(\pm\) SE across 99 OptFeynman problems. Lower values indicate closer distribution recovery.} \vspace{2mm}
\label{tab:optfeynman-continuous-kl}
\small
\setlength{\tabcolsep}{6pt}
\renewcommand{\arraystretch}{1.12}
\begin{tabular}{@{}llcc@{}}
\toprule
Condition & Representation & KL & Median [Q1,Q3] \\
\midrule
Clean
& OptScientist law
& \(0.0461\pm0.0111\)
& \(0.0080\,[0.0013,0.0358]\) \\
Clean
& OptLearner score
& \(0.0078\pm0.0032\)
& \(0.0015\,[0.0010,0.0030]\) \\
Noisy-context
& OptScientist law
& \(0.0460\pm0.0163\)
& \(0.0071\,[0.0012,0.0382]\) \\
Noisy-context
& OptLearner score
& \(0.0082\pm0.0036\)
& \(0.0014\,[0.0009,0.0031]\) \\
\bottomrule
\end{tabular}
\end{table}

For every value of \(K\), the median and both quartiles of OptScientist exact-order accuracy are equal to one. Most problems therefore retain perfect ordering even when the candidate set becomes substantially denser. Pairwise recovery is similarly stable. Mean OptScientist pairwise accuracy remains approximately \(0.96\) across all candidate-set sizes and both training conditions, with a median of one.

OptLearner also achieves high pairwise accuracy, while its exact-order accuracy decreases more strongly with \(K\). Under the Clean condition, mean OptLearner exact-order accuracy decreases from \(0.825\) at \(K=10\) to \(0.719\) at \(K=100\).

The recovered continuous laws from OptScientist also preserve the Gibbs distributions generated by the ground-truth objectives. Mean dense-grid KL divergence is \(0.0461\) under Clean and \(0.0460\) under Noisy-context. The corresponding medians are \(0.0080\) and \(0.0071\), showing low distributional error for most tasks. OptLearner achieves mean KL values of \(0.0078\) and \(0.0082\), consistent with its direct fit to the stochastic optimization landscape.

\textbf{Discrete outcomes.}
The discrete results show a similar pattern. Under the Clean condition, the recovered OptScientist laws achieve mean exact-order accuracy \(0.971\) and pairwise accuracy \(0.984\). Under Noisy-context, these values are \(0.974\) and \(0.984\). The median and both quartiles of OptScientist exact-order accuracy are equal to one in both conditions.

\begin{table}[htbp]
\centering
\caption{\textbf{Preference recovery for discrete OptFeynman problems.}
Values are mean \(\pm\) SE across 99 problems. Median and interquartile range refer to exact-order accuracy.} \vspace{2mm}
\label{tab:optfeynman-discrete}
\footnotesize
\setlength{\tabcolsep}{3.8pt}
\renewcommand{\arraystretch}{1.12}
\begin{tabular}{@{}llcccc@{}}
\toprule
Condition & Representation
& Exact order
& Median [Q1,Q3]
& Pairwise
& Exact \(=1\) \\
\midrule
Clean
& OptScientist law
& \(0.971\pm0.014\)
& \(1.000\,[1.000,1.000]\)
& \(0.984\pm0.011\)
& \(88/99\) \\
Clean
& OptLearner score
& \(0.805\pm0.025\)
& \(0.920\,[0.675,1.000]\)
& \(0.947\pm0.009\)
& \(31/99\) \\
Noisy-context
& OptScientist law
& \(0.974\pm0.015\)
& \(1.000\,[1.000,1.000]\)
& \(0.984\pm0.011\)
& \(92/99\) \\
Noisy-context
& OptLearner score
& \(0.805\pm0.024\)
& \(0.890\,[0.705,1.000]\)
& \(0.948\pm0.009\)
& \(27/99\) \\
\bottomrule
\end{tabular}
\end{table}

Training-context noise produces little change in the overall recovery statistics. Mean OptScientist exact-order accuracy changes from \(0.971\) to \(0.974\), while mean pairwise accuracy remains approximately \(0.984\). The number of perfectly recovered tasks changes from \(88/99\) to \(92/99\). These results are consistent with the robustness observed in
Figure~\ref{fig:optfeynman-noisyx}.

\begin{table}[htbp]
\centering
\caption{\textbf{Distribution recovery for discrete OptFeynman problems.}
The ground-truth Gibbs distribution over five candidate actions is compared with the distribution induced by the recovered score. Probability MAE is the mean absolute error between ground-truth and recovered Gibbs probabilities. Values are mean \(\pm\) SE across 99 problems. Lower values indicate closer recovery.}
\label{tab:optfeynman-discrete-probability} \vspace{2mm}
\scriptsize
\setlength{\tabcolsep}{1.8pt}
\renewcommand{\arraystretch}{1.08}
\begin{tabular}{@{}llcccc@{}}
\toprule
Condition & Representation
& Prob. MAE
& MAE median [Q1,Q3]
& KL
& KL median [Q1,Q3] \\
\midrule
Clean
& OptScientist law
& \(0.0182\pm0.0025\)
& \(0.0115\,[0.0061,0.0201]\)
& \(0.0217\pm0.0092\)
& \(0.0036\,[0.0008,0.0109]\) \\
Clean
& OptLearner score
& \(0.0161\pm0.0032\)
& \(0.0112\,[0.0083,0.0154]\)
& \(0.1932\pm0.1873\)
& \(0.0032\,[0.0016,0.0059]\) \\
Noisy-context
& OptScientist law
& \(0.0208\pm0.0036\)
& \(0.0117\,[0.0062,0.0238]\)
& \(0.2005\pm0.1852\)
& \(0.0036\,[0.0010,0.0132]\) \\
Noisy-context
& OptLearner score
& \(0.0161\pm0.0032\)
& \(0.0110\,[0.0080,0.0146]\)
& \(0.2025\pm0.1964\)
& \(0.0033\,[0.0016,0.0052]\) \\
\bottomrule
\end{tabular}
\end{table}

For the Gibbs distributions over the five candidate actions, we additionally report probability MAE, defined as the mean absolute error between the ground-truth and recovered Gibbs probabilities across candidate actions. The OptScientist laws achieve mean probability MAE \(0.0182\) and mean KL divergence \(0.0217\) under the Clean condition. Their medians are \(0.0115\) and \(0.0036\), respectively. Under Noisy-context, mean probability MAE remains low at \(0.0208\), with a median of \(0.0117\). Mean OptScientist KL increases to \(0.2005\), while the median remains \(0.0036\) and the third quartile is \(0.0132\). The difference between the mean and median reflects a small number of tasks with large KL values.

\begin{table}[htbp]
\centering
\normalsize
\setlength{\tabcolsep}{6pt}
\renewcommand{\arraystretch}{1.6}
\caption{
\textbf{Examples of recovered OptFeynman laws.}
Ground-truth and recovered laws are shown in optimization form after removing
numerical constants and scale factors that do not affect the optimizer.
} \vspace{2mm}
\label{tab:optfeynman-recovered-examples}
\begin{tabular}{@{}p{0.20\textwidth}p{0.35\textwidth}p{0.35\textwidth}@{}}
\toprule
Decision variable & Effective ground truth & Effective recovered expression \\
\midrule
\(y_3\)
& \(\displaystyle \max_{y_3}\; x_3 y_3\)
& \(\displaystyle \max_{y_3}\; x_3 y_3\) \\
\addlinespace[0.35em]

\(w\)
& \(\displaystyle \max_{w}\; w^2\)
& \(\displaystyle \max_{w}\; w^2\) \\
\addlinespace[0.35em]

\(\chi\)
& \(\displaystyle \max_{\chi}\; B\chi\)
& \(\displaystyle \max_{\chi}\; B\chi\) \\
\addlinespace[0.35em]

\(E_f\)
& \(\displaystyle \max_{E_f}\; E_f\cos\theta\)
& \(\displaystyle \max_{E_f}\; E_f\cos\theta\) \\
\addlinespace[0.35em]

\(\theta\)
& \(\displaystyle \max_{\theta}\; \sin\theta\)
& \(\displaystyle \max_{\theta}\; \sin\theta\) \\
\addlinespace[0.35em]

\(G\)
& \(\displaystyle \max_{G}\; G\left(\frac{1}{r_2}-\frac{1}{r_1}\right)\)
& \(\displaystyle \max_{G}\; \frac{G(r_1-r_2)}{r_1r_2}\) \\
\addlinespace[0.35em]

\(x\)
& \(\displaystyle \max_{x}\; x^2\)
& \(\displaystyle \max_{x}\; x^2\) \\
\addlinespace[0.35em]

\(d\)
& \(\displaystyle \max_{d}\; \frac{T_2-T_1}{d}\)
& \(\displaystyle \max_{d}\; \frac{T_2-T_1}{d}\) \\
\addlinespace[0.35em]

\(r\)
& \(\displaystyle \max_{r}\; \frac{1}{r^2}\)
& \(\displaystyle \max_{r}\; \frac{1}{r^2}\) \\
\addlinespace[0.35em]

\(d\)
& \(\displaystyle \max_{d}\; \frac{1}{d}\)
& \(\displaystyle \max_{d}\; \frac{1}{d}\) \\
\bottomrule
\end{tabular}
\end{table}

\textbf{Effect of OptDiscover complexity control.}
Here we evaluate the expression-level reduction used within the symbolic complexity control of OptDiscover described in Sec.~\ref{sec:optdiscover}. In Table~\ref{tab:optfeynman-pruning}, \emph{Initial} denotes the recovered symbolic expression before this reduction, and \emph{Final} denotes the compact symbolic expression after it. We additionally compare exact-order recovery before and after this reduction. Discrete recovery is unchanged: mean exact-order accuracy remains \(0.971\) under Clean and \(0.974\) under Noisy-context, with \(88\) and \(92\) perfectly recovered problems, respectively.

For continuous outcomes, complexity control gives small improvements in exact-order recovery. Under Clean, mean accuracy increases by \(0.010\), \(0.011\), and \(0.013\) at \(K=10\), \(50\), and \(100\). At \(K=100\), the number of perfectly recovered problems increases from \(67\) to \(78\). Under Noisy-context, changes in mean accuracy range from \(-0.001\) to \(+0.005\), and perfect recovery at \(K=100\) increases from \(70\) to \(76\) problems.

\begin{table}[htbp]
\centering
\caption{\textbf{Exact-order recovery before and after OptDiscover complexity control.}
Mean accuracy is computed across matched OptFeynman problems. Initial denotes the recovered symbolic expression before expression-level reduction, and Final denotes the compact symbolic expression after it. The final column reports the number of problems with perfect exact-order recovery before and after this reduction.}
\label{tab:optfeynman-pruning}
\footnotesize
\setlength{\tabcolsep}{3.8pt}
\renewcommand{\arraystretch}{1.12} \vspace{2mm}
\begin{tabular}{@{}llcccc@{}}
\toprule
Decision type & Condition
& Initial
& Final
& \(\Delta\)
& Perfect recovery \\
\midrule
Continuous
& Clean, \(K=10\)
& \(0.890\)
& \(0.900\)
& \(+0.010\)
& \(77\rightarrow78\) \\
Continuous
& Clean, \(K=50\)
& \(0.860\)
& \(0.871\)
& \(+0.011\)
& \(72\rightarrow78\) \\
Continuous
& Clean, \(K=100\)
& \(0.852\)
& \(0.866\)
& \(+0.013\)
& \(67\rightarrow78\) \\
Continuous
& Noisy-context, \(K=10\)
& \(0.902\)
& \(0.902\)
& \(-0.001\)
& \(76\rightarrow76\) \\
Continuous
& Noisy-context, \(K=50\)
& \(0.872\)
& \(0.874\)
& \(+0.002\)
& \(75\rightarrow76\) \\
Continuous
& Noisy-context, \(K=100\)
& \(0.864\)
& \(0.869\)
& \(+0.005\)
& \(70\rightarrow76\) \\
\midrule
Discrete
& Clean
& \(0.971\)
& \(0.971\)
& \(0.000\)
& \(88\rightarrow88\) \\
Discrete
& Noisy-context
& \(0.974\)
& \(0.974\)
& \(0.000\)
& \(92\rightarrow92\) \\
\bottomrule
\end{tabular}
\end{table}

\begin{table}[t]
\centering
\caption{\textbf{Held-out baseline recovery for discrete OptFeynman problems.}
Exact-order and pairwise preference accuracy are evaluated on unseen test
contexts. Exact \(=1\) reports the percentage of tasks with perfect exact-order
recovery. KL divergence compares the ground-truth and recovered Gibbs
distributions over the five candidate actions; lower values indicate closer
distribution recovery. For discrete decisions, a standard multiclass MLP can be
written in the same finite-action softmax form as a conditional Neural EBM, so
we use the multiclass MLP directly rather than introducing a separate Neural EBM
baseline.}
\vspace{2mm}
\label{tab:optfeynman-baselines-discrete}
\footnotesize
\setlength{\tabcolsep}{3.8pt}
\renewcommand{\arraystretch}{1.12}
\begin{tabular}{@{}llrcccc@{}}
\toprule
Condition & Method & \(n\)
& Exact order
& Exact \(=1\)
& Pairwise
& KL \\
\midrule
Clean
& GBDT
& 99 & 0.327 & 0.0\% & 0.844 & 0.1679 \\

Clean
& LR
& 99 & 0.716 & 17.2\% & 0.919 & 0.0859 \\

Clean
& MLP
& 99 & 0.420 & 6.1\% & 0.859 & 0.1580 \\

Clean
& RF
& 99 & 0.206 & 0.0\% & 0.794 & 0.5992 \\

Clean
& XGBoost
& 99 & 0.337 & 0.0\% & 0.847 & 0.0833 \\

Clean
& SR
& 99 & 0.076 & 0.0\% & 0.743 & 0.8855 \\

Clean
& LLM-only
& 99 & 0.588 & 5.1\% & 0.911 & 0.0125 \\

Clean
& GeCCo
& 99 & 0.715 & 20.2\% & 0.928 & 0.2025 \\

Clean
& MDA
& 99 & 0.700 & 21.2\% & 0.926 & 0.1579 \\

Clean
& BL
& 99 & 0.805 & 31.3\% & 0.947 & 0.1932 \\

Clean
& OptScientist
& 99 & 0.971 & 88.9\% & 0.984 & 0.0217 \\
\midrule
Noisy-context
& GBDT
& 99 & 0.371 & 0.0\% & 0.854 & 0.1744 \\

Noisy-context
& LR
& 99 & 0.717 & 17.2\% & 0.919 & 0.0860 \\

Noisy-context
& MLP
& 99 & 0.418 & 6.1\% & 0.860 & 0.1635 \\

Noisy-context
& RF
& 99 & 0.104 & 0.0\% & 0.701 & 1.1334 \\

Noisy-context
& XGBoost
& 99 & 0.331 & 0.0\% & 0.842 & 0.0636 \\

Noisy-context
& SR
& 99 & 0.076 & 1.0\% & 0.743 & 0.8816 \\

Noisy-context
& LLM-only
& 99 & 0.593 & 8.1\% & 0.911 & 0.1214 \\

Noisy-context
& GeCCo
& 99 & 0.721 & 20.2\% & 0.930 & 0.1592 \\

Noisy-context
& MDA
& 99 & 0.703 & 18.2\% & 0.930 & 0.1417 \\

Noisy-context
& BL
& 99 & 0.805 & 27.3\% & 0.948 & 0.2025 \\

Noisy-context
& OptScientist
& 99 & 0.974 & 92.9\% & 0.984 & 0.2005 \\
\bottomrule
\end{tabular}
\end{table}

\begin{table}[htbp]
\centering
\caption{\textbf{Held-out baseline recovery for continuous OptFeynman problems.}
Within the exact-order, perfect-recovery and pairwise columns, entries correspond
to \(K=10/50/100\), respectively. Exact \(=1\) reports the percentage of tasks
with perfect exact-order recovery. KL divergence compares the ground-truth and
recovered Gibbs distributions on the dense held-out test grid; lower values
indicate closer distribution recovery.}
\vspace{2mm}
\label{tab:optfeynman-baselines-continuous}
\scriptsize
\setlength{\tabcolsep}{2.5pt}
\renewcommand{\arraystretch}{1.14}
\begin{tabular}{@{}llrcccc@{}}
\toprule
Condition & Method & \(n\)
& \(\text{Exact order}_{10/50/100}\)
& \(\text{Exact}=1~(\%)_{10/50/100}\)
& \(\text{Pairwise}_{10/50/100}\)
& KL \\
\midrule
Clean
& GBDT
& 99
& \(0.063/0.003/0.002\)
& \(0.0/0.0/0.0\)
& \(0.704/0.704/0.704\)
& 0.6362 \\

Clean
& Neural EBM
& 99
& \(0.482/0.314/0.294\)
& \(12.1/10.1/10.1\)
& \(0.888/0.887/0.888\)
& 0.0098 \\

Clean
& MLP
& 99
& \(0.061/0.003/0.001\)
& \(0.0/0.0/0.0\)
& \(0.703/0.703/0.703\)
& 0.6349 \\

Clean
& RF
& 99
& \(0.064/0.003/0.001\)
& \(0.0/0.0/0.0\)
& \(0.703/0.702/0.703\)
& 0.6408 \\

Clean
& XGBoost
& 99
& \(0.061/0.003/0.001\)
& \(0.0/0.0/0.0\)
& \(0.704/0.704/0.704\)
& 0.6399 \\

Clean
& SR
& 99
& \(0.060/0.002/0.001\)
& \(0.0/0.0/0.0\)
& \(0.710/0.710/0.711\)
& 0.6414 \\

Clean
& LLM-only
& 99
& \(0.061/0.003/0.002\)
& \(0.0/0.0/0.0\)
& \(0.701/0.702/0.702\)
& 0.6422 \\

Clean
& GeCCo
& 99
& \(0.062/0.003/0.002\)
& \(0.0/0.0/0.0\)
& \(0.703/0.704/0.704\)
& 0.6353 \\

Clean
& MDA
& 99
& \(0.062/0.002/0.002\)
& \(0.0/0.0/0.0\)
& \(0.704/0.704/0.704\)
& 0.6355 \\

Clean
& BL
& 99
& \(0.825/0.743/0.719\)
& \(30.3/20.2/14.1\)
& \(0.960/0.960/0.960\)
& 0.0078 \\

Clean
& OptScientist
& 99
& \(0.900/0.871/0.866\)
& \(78.8/78.8/78.8\)
& \(0.963/0.963/0.963\)
& 0.0461 \\
\midrule
Noisy-context
& GBDT
& 99
& \(0.063/0.003/0.001\)
& \(0.0/0.0/0.0\)
& \(0.704/0.704/0.705\)
& 0.6372 \\

Noisy-context
& Neural EBM
& 99
& \(0.473/0.307/0.290\)
& \(10.1/6.1/6.1\)
& \(0.887/0.887/0.888\)
& 0.0101 \\

Noisy-context
& MLP
& 99
& \(0.061/0.003/0.001\)
& \(0.0/0.0/0.0\)
& \(0.703/0.703/0.703\)
& 0.6351 \\

Noisy-context
& RF
& 99
& \(0.081/0.006/0.002\)
& \(0.0/0.0/0.0\)
& \(0.675/0.674/0.674\)
& 0.8009 \\

Noisy-context
& XGBoost
& 99
& \(0.062/0.003/0.002\)
& \(0.0/0.0/0.0\)
& \(0.704/0.704/0.704\)
& 0.6399 \\

Noisy-context
& SR
& 99
& \(0.061/0.002/0.001\)
& \(0.0/0.0/0.0\)
& \(0.710/0.710/0.710\)
& 0.6437 \\

Noisy-context
& LLM-only
& 99
& \(0.062/0.003/0.002\)
& \(0.0/0.0/0.0\)
& \(0.704/0.704/0.704\)
& 0.6388 \\

Noisy-context
& GeCCo
& 99
& \(0.062/0.003/0.002\)
& \(0.0/0.0/0.0\)
& \(0.704/0.704/0.704\)
& 0.6353 \\

Noisy-context
& MDA
& 99
& \(0.062/0.003/0.002\)
& \(0.0/0.0/0.0\)
& \(0.703/0.704/0.704\)
& 0.6358 \\

Noisy-context
& BL
& 99
& \(0.826/0.743/0.722\)
& \(30.3/23.2/15.2\)
& \(0.960/0.960/0.960\)
& 0.0082 \\

Noisy-context
& OptScientist
& 99
& \(0.902/0.874/0.869\)
& \(76.8/76.8/76.8\)
& \(0.959/0.959/0.958\)
& 0.0460 \\
\bottomrule
\end{tabular}
\end{table}

\textbf{Held-out baseline comparison.}
Tables~\ref{tab:optfeynman-baselines-discrete} and
\ref{tab:optfeynman-baselines-continuous} report ground-truth recovery for the
baseline models on unseen test contexts. The largest separation appears in
exact-order recovery, which requires the recovered score to preserve the complete
preference order induced by the ground-truth objective. In the discrete Clean
setting, the two strongest baselines by mean exact-order accuracy are BL and
logistic regression. BL achieves mean exact-order accuracy \(0.805\) with perfect
recovery on \(31.3\%\) of tasks, and logistic regression achieves \(0.716\) with
perfect recovery on \(17.2\%\). OptScientist reaches \(0.971\) exact-order
accuracy and perfectly recovers \(88/99\) tasks. The separation becomes larger
for continuous outcomes as the candidate set grows. At \(K=100\), the two
strongest baselines by mean exact-order accuracy, BL and Neural EBM, achieve
exact-order accuracies of \(0.719\) and \(0.294\), with perfect recovery on
\(14.1\%\) and \(10.1\%\) of tasks, respectively. OptScientist retains
exact-order accuracy \(0.866\) and perfectly recovers \(78/99\) tasks.

Pairwise and distributional metrics provide complementary views of recovery.
Several baselines preserve a substantial fraction of pairwise comparisons or
closely approximate the Gibbs distribution while showing considerably weaker
recovery of the complete ordering. For example, Neural EBM reaches pairwise
accuracy \(0.888\) and dense-grid KL divergence \(0.0098\) at \(K=100\) in the
continuous Clean setting, while its exact-order accuracy is \(0.294\). BL
achieves a dense-grid KL divergence of \(0.0078\), lower than that achieved by
OptScientist, while its exact-order accuracy decreases to \(0.719\) at
\(K=100\). A similar pattern appears for the LLM-only baseline in the discrete
Clean setting, which reaches KL divergence \(0.0125\) but exact-order accuracy
\(0.588\), with perfect recovery on only \(5.1\%\) of tasks. These results show
that accurate approximation of local pairwise relations or the stochastic
landscape can coexist with substantial errors in the complete preference order.
Because all preference metrics are evaluated on held-out contexts, exact-order
recovery places a stronger requirement on the learned model: the recovered
mechanism must preserve the full ordering generated by the ground-truth
objective across unseen contexts.

The LLM-based model-discovery baselines show particularly limited recovery in
the continuous setting. On discrete Clean OptFeynman, GeCCo and MDA achieve
exact-order accuracies of \(0.715\) and \(0.700\), respectively, remaining below
BL and substantially below OptScientist. On continuous Clean OptFeynman, their
performance decreases sharply as the candidate set becomes denser. At
\(K=100\), LLM-only, GeCCo and MDA achieve exact-order accuracies of
\(0.0018\), \(0.0021\), and \(0.0017\), respectively, with no perfectly
recovered tasks; their pairwise accuracies remain around \(0.70\), and their
dense-grid KL divergences remain around \(0.64\). A plausible interpretation is
that LLM-based discovery depends strongly on the scientific and functional-form
priors available in language. When such priors provide limited information about
the generating mechanism, identifying a transferable law directly from
observations becomes substantially more difficult. Strong textual priors may
also favour familiar or plausible functional forms, creating a risk that the
discovered expression reflects prior knowledge more strongly than the mechanism
that generated the observations. The OptFeynman results therefore highlight the
value of learning the optimization landscape from data before symbolic
distillation, particularly when informative domain priors are unavailable or
incomplete.

\textbf{LLM baseline protocol.}
For the LLM-based baselines on OptFeynman, model generation, revision, and selection use only the training and validation splits. All methods use the same DeepSeek-V4-Flash model and follow their original discovery procedures with a unified budget of eight discovery rounds. To prevent semantic leakage from physics-specific notation, all OptFeynman variable names are replaced with neutral identifiers before being provided to the LLMs. The held-out test set is never exposed during model discovery or model selection. After the final model is selected, its predictive model is evaluated directly on the test set using the common OptFeynman metrics; this final evaluation is fully deterministic and involves no further LLM interaction.

Overall, the detailed statistics support the ground-truth recovery results reported in the main text. OptScientist recovers the complete preference order on most OptFeynman problems, with high pairwise agreement across the benchmark. It substantially outperforms the baseline methods in exact-order recovery, particularly in the fraction of problems for which the complete ground-truth preference order is perfectly recovered. Continuous recovery remains stable as the candidate set increases from \(K=10\) to \(K=100\), and recovery remains strong when Gaussian noise is added to the training contexts. The problem-level medians and quartiles show that these results hold across most of the benchmark optimization problems.

\subsection{Human Choice under Risk and Ambiguity}
\label{sec:add-cpc18}

\subsubsection{Dataset and evaluation}
\label{sec:add-cpc18-dataset}

The CPC18 experiment studies repeated binary choices between two monetary prospects under risk and ambiguity \citep{plonsky2019cpc18}. We use a random split with
\(408{,}600\), \(51{,}060\) and \(51{,}090\) observations for training,
validation and testing, respectively. For Figure~\ref{fig:cpc18}, test metrics
are first computed at the participant level and then averaged across the \(686\)
test participants. Error bars denote the standard error of the mean across
participant-level metrics.

\subsubsection{Theory Prior Agent and variable discovery}
\label{sec:add-cpc18-prior-analysis}

Before symbolic discovery, the Theory Prior Agent used scientific theories of
decision-making under risk, ambiguity and repeated feedback to construct
theoretically motivated variables beyond those directly available in the CPC18
data. These variables were specified independently of validation and test
performance and were subsequently considered as candidate inputs for symbolic
discovery.

Let option \(j\in\{A,B\}\) have high payoff \(H_j\), low payoff \(L_j\), and
probability \(p_{H,j}\) of the high payoff. The agent constructed expected-value
representations
\[
\mathrm{EV}_A
=
p_{H,A}H_A+(1-p_{H,A})L_A,
\qquad
\mathrm{EV}_B
=
p_{H,B}H_B+(1-p_{H,B})L_B,
\]
together with relative probability and payoff-dispersion measures,
\[
\Delta p_H=p_{H,B}-p_{H,A},
\qquad
\Delta \mathrm{Spread}
=
(H_B-L_B)-(H_A-L_A).
\]
It also introduced an ambiguity--correlation interaction,
\[
\mathrm{AmbNegCorr}
=
\mathrm{Amb}\,
\mathbbm{1}\{\mathrm{Corr}<0\},
\]
and an indicator for the feedback framing condition,
\[
\mathrm{ByFB}
=
\mathbbm{1}\{\mathrm{Condition}=\texttt{ByFB}\}.
\]

Participant-level variables included gender,
\[
\mathrm{Male}
=
\mathbbm{1}\{\mathrm{Gender}=\mathrm{M}\},
\]
and an indicator for residence in wealthier regions,
\[
\mathrm{WealthyRegion}
=
\mathbbm{1}\{\mathrm{Region}\in\mathcal{W}\},
\]
where \(\mathcal{W}\) denotes the corresponding region classification used in
preprocessing.

Because CPC18 contains repeated choices with feedback, the Theory Prior Agent
also proposed lagged obtained and forgone payoff variables. Let \(R_t^A\) and
\(R_t^B\) denote the payoff feedback associated with options \(A\) and \(B\),
and let \(y_t\in\{0,1\}\) indicate whether option \(B\) was chosen. The obtained
and forgone payoffs are
\[
R_t^{\mathrm{obt}}
=
(1-y_t)R_t^A+y_tR_t^B,
\qquad
R_t^{\mathrm{forgone}}
=
y_tR_t^A+(1-y_t)R_t^B.
\]
For prediction at round \(t\), only lagged quantities from preceding rounds were
used. During symbolic discovery, OptScientist Model subsequently identified
decision-relevant combinations of these variables, including the
counterfactual payoff difference
\[
\Delta R_t^{\mathrm{cf}}
=
R_t^{\mathrm{forgone}}-R_t^{\mathrm{obt}}.
\]

Together, the Theory Prior Agent provides candidate representations of expected
reward, probability sensitivity, risk, ambiguity, framing, demographic
heterogeneity and feedback, while the OptScientist Model determines how these
variables enter the discovered symbolic law.

\subsubsection{Direct symbolic law}
\label{sec:add-cpc18-direct-law}

For the binary-choice task presented in the CPC18 dataset, let \(y_t=1\) denote choosing option B and
\(y_t=0\) denote choosing option A. The direct OptScientist law can be written as
\begin{equation}
    \widehat U(x_t,y_t)=y_t D(x_t),
    \qquad
    \widehat y_t
    =
    \arg\max_{y_t\in\{0,1\}}
    \widehat U(x_t,y_t),
    \label{eq:add-cpc18-direct-utility}
\end{equation}
where \(D(x_t)\) is the recovered score difference between choosing B and
choosing A. The selected expression is
\begin{align}
D(x_t)= {}&
-0.8173\,EV_A
+0.5701\,EV_B
+0.3393\,L_B
-0.0709\,\mathrm{PrevForgone} \notag\\
& -0.8173\,\mathrm{PrevExp}
+1.6541\,\mathrm{PrevChoice}_B
+0.0709\,\mathrm{PrevPayoff}.
\label{eq:add-cpc18-direct-score-diff}
\end{align}
The model predicts that people would choose option B when \(D(x_t)>0\). Continuous variables are standardized
using the training split. The expression has symbolic complexity 23 and
contains five distinct numerical constants.

Equation~\eqref{eq:add-cpc18-direct-score-diff} supports the abstract law shown in
Figure~\ref{fig:cpc18}b. The current lottery terms form the immediate
reward component \(V(x_t,y_t)\). The previous-choice term captures persistence
in the last action, corresponding to the switching-cost component. The previous
payoff and forgone-payoff terms capture sequential feedback from the previous
round.

\subsubsection{Automatic-feature symbolic law}
\label{sec:add-cpc18-autof-law}

AutoF OptScientist follows the automatic-feature extraction route introduced in
the OptScientist Model. It first maps \((x_t,y_t)\) into an intermediate
representation,
\begin{equation}
    h(x_t,y_t)=(h_1,\ldots,h_8),
\end{equation}
taken from an OptLearner inner layer. OptDiscover then distills the final score
as a symbolic function of these learned features:
\begin{align}
s(h)={}&
0.04345\,h_1
+0.00976\,h_2
+0.11970\,h_3
+0.04345\,h_6
+0.15760\,h_7 \notag\\
&+0.04345\,|h_6|
+0.11970\,\left|
0.1819\,h_3
-0.1819\,h_4
+0.3752\,h_7
\right|
-1.3408 .
\label{eq:add-cpc18-inner-feature-score}
\end{align}
The induced binary score difference is
\begin{equation}
    D_{\mathrm{FM}}(x_t)
    =
    s\!\left(h(x_t,1)\right)
    -
    s\!\left(h(x_t,0)\right).
    \label{eq:add-cpc18-feature-mediated-diff}
\end{equation}
The model predicts that people would choose option B when \(D_{\mathrm{FM}}(x_t)>0\). This expression has outer
symbolic complexity 33 and contains seven distinct numerical constants. This
automatic-feature route allows OptLearner to compress weakly structured
behavioural information before symbolic extraction.

\subsubsection{Predictive performance and complexity}
\label{sec:add-cpc18-performance}

\begin{table}[htbp]
\centering
\scriptsize
\setlength{\tabcolsep}{2pt}
\renewcommand{\arraystretch}{1.15}
\caption{
\textbf{Predictive performance and complexity on CPC18.}
Metrics are reported as participant-level mean \(\pm\) standard error across test
participants. Parameter count denotes the number of distinct numerical constants in
the recovered symbolic law.
} \vspace{2mm}
\label{tab:add-cpc18-performance}
\begin{tabular}{@{}lccccccc@{}}
\toprule
Variant & Parameters & Accuracy$\uparrow$ & AUC$\uparrow$ & Prob. MSE$\downarrow$ & NLL$\downarrow$ & KL$\downarrow$ & ECE$\downarrow$ \\
\midrule
Direct OptScientist
& 5
& \(0.829\pm0.004\)
& \(0.863\pm0.004\)
& \(0.149\pm0.002\)
& \(0.476\pm0.004\)
& \(0.011\pm0.001\)
& \(0.172\pm0.002\) \\

AutoF OptScientist
& 7
& \(0.798\pm0.004\)
& \(0.865\pm0.004\)
& \(0.137\pm0.002\)
& \(0.422\pm0.006\)
& \(0.006\pm0.000\)
& \(0.116\pm0.001\) \\
\bottomrule
\end{tabular}
\end{table}

Table~\ref{tab:add-cpc18-performance} reports participant-level predictive
performance and parameter counts for the two OptScientist variants. Direct
OptScientist achieves higher choice accuracy with fewer parameters, whereas
AutoF OptScientist yields slightly higher AUC and better probabilistic
calibration, with lower probability MSE, NLL, KL divergence and ECE. Thus,
direct symbolic discovery provides the more parsimonious decision rule, while
automatic feature extraction improves the fidelity of predicted choice
probabilities.

Notably, the MLP baseline achieves approximately 84\% held-out accuracy.
Together with the clustering of the strongest predictive models around
82--84\%, this suggests that predictive performance on CPC18 is already close
to saturation.

\subsection{Central-Bank Monetary Policy}
\label{sec:add-central-bank}

\subsubsection{Dataset and evaluation}
\label{sec:add-central-bank-dataset}

The central-bank experiment studies quarterly monetary-policy decisions from
macro-financial conditions. The dataset contains \(127\) quarterly observations.
We use \(90\), \(23\) and \(14\) observations for training, validation and
out-of-time testing, respectively. The outcome is a five-class monetary policy stance
\(y_t\in\{-2,-1,0,1,2\}\), ranging from large easing to large tightening.

Continuous model predictions are mapped to the observed five policy classes
before computing exact policy-stance accuracy. Policy-direction accuracy
compares the sign of the mapped policy class, distinguishing easing, no change
and tightening. Error bars in Figure~\ref{fig:central-bank} are bootstrap
standard errors from \(10{,}000\) resamples of the \(14\) test quarters with
replacement.

\subsubsection{Single-\(U\) symbolic law}
\label{sec:add-central-bank-single-u}

The single-\(U\) OptScientist Model distills the learned policy landscape into
one symbolic objective over the candidate policy stance \(y_t\). The concrete
distilled symbolic branch is
\begin{align}
U_t(y) = y\Big(&
0.01812\,a_t r_{t-1}
-0.01812\,a_t \Delta_t
+0.01812\,a_t \ell_t
+0.01812\,a_t y \notag\\
&-0.33822\,c_t
+0.19520\,g_t
-0.04029\,r_{t-1}\Delta_t
+0.01812\,r_{t-1}\ell_t
+0.01812\,r_{t-1}u_t \notag\\
&+0.04029\,r_{t-1}y
-0.27460\,r_{t-1}
+0.04029\,\Delta_t^2
-0.05841\,\Delta_t \ell_t
-0.01812\,\Delta_t u_t \notag\\
&-0.08058\,\Delta_t y
+0.93769\,\Delta_t
+0.01812\,\ell_t^2
+0.01812\,\ell_t u_t
+0.05841\,\ell_t y \notag\\
&-0.46980\,\ell_t
+0.19520\,\tau_t
+0.01812\,u_t y
+0.04029\,y^2
-0.93769\,y
\Big),
\label{eq:add-central-bank-single-u}
\end{align}
with final score
\begin{equation}
    S_t(y)=0.58028\,U_t(y)+0.72030 .
    \label{eq:add-central-bank-single-u-score}
\end{equation}
Here \(a_t\) denotes the adjusted national financial conditions index,
\(r_{t-1}\) the previous average target rate, \(\Delta_t\) the previous
quarterly change in the target rate, \(\ell_t\) the St. Louis financial stress
index, \(c_t\) the credit spread, \(g_t\) real GDP growth, \(u_t\) unemployment
and \(\tau_t\) the term spread. All continuous variables are standardized using
the training split. The single-\(U\) expression has final effective symbolic
complexity \(131\) under the decision-relevant complexity accounting used in
the experiment.

\subsubsection{Rewriting the symbolic law into target-tracking form}
\label{sec:add-central-bank-target-rewrite}

The target-tracking form in Figure~\ref{fig:central-bank}a is obtained by
grouping the decision-relevant terms of the discovered score. Since
\(0.58028>0\), the positive head scale in Eq.~\eqref{eq:add-central-bank-single-u-score}
preserves the maximizer over \(y\). The head bias is constant across candidate
policy actions and is removed from the decision-relevant expression. After this
normalization, Eq.~\eqref{eq:add-central-bank-single-u} can be collected as a
polynomial in the candidate policy action:
\begin{equation}
    S_t(y)
    \;\sim\;
    A_t y + B_t y^2 + d y^3 ,
    \label{eq:add-central-bank-polynomial-score}
\end{equation}
where \(\sim\) denotes equivalence up to positive scaling and additive
context-only constants. The linear policy-pressure term is
\begin{align}
A_t={}&
0.01812\,a_t r_{t-1}
-0.01812\,a_t \Delta_t
+0.01812\,a_t \ell_t
-0.33822\,c_t
+0.19520\,g_t \notag\\
&-0.04029\,r_{t-1}\Delta_t
+0.01812\,r_{t-1}\ell_t
+0.01812\,r_{t-1}u_t
-0.27460\,r_{t-1} \notag\\
&+0.04029\,\Delta_t^2
-0.05841\,\Delta_t\ell_t
-0.01812\,\Delta_t u_t
+0.93769\,\Delta_t \notag\\
&+0.01812\,\ell_t^2
+0.01812\,\ell_t u_t
-0.46980\,\ell_t
+0.19520\,\tau_t .
\label{eq:add-central-bank-linear-pressure}
\end{align}
The quadratic and cubic decision-residual terms are
\begin{align}
B_t={}&
0.01812\,a_t
+0.04029\,r_{t-1}
-0.08058\,\Delta_t
+0.05841\,\ell_t
+0.01812\,u_t
-0.93769,
\label{eq:add-central-bank-quadratic-term}\\
d={}&0.04029 .
\label{eq:add-central-bank-cubic-term}
\end{align}

For better interpretability, the Scientific Critic Agent abstracts the leading linear term
\(A_t\) into four macro-policy pressures,
\begin{equation}
    A_t \leadsto N_t
    =
    m(g_t+\tau_t)
    -m(f_t+q c_t)
    -\rho r_{t-1}
    +(s+\rho)\Delta_t .
    \label{eq:add-central-bank-abstract-pressure}
\end{equation}
Here \(g_t+\tau_t\) groups real activity and the term spread as macroeconomic
strength; \(f_t+q c_t\) groups financial stress and credit conditions as
financial fragility; \(r_{t-1}\) summarizes inherited policy tightness; and
\(\Delta_t\) summarizes policy momentum. Remaining interaction terms,
state-dependent curvature and higher-order terms are retained in the residual
decision term below.

Writing \(B_t=-\kappa+\delta_t\) with \(\kappa>0\), the decision-relevant score of Eq.~\ref{eq:add-central-bank-polynomial-score}
can be rewritten as
\begin{equation}
    S_t(y)
    \;\sim\;
    N_t y-\kappa y^2
    +\delta_t y^2
    +d y^3 .
    \label{eq:add-central-bank-target-score}
\end{equation}
Maximizing this score is equivalent to minimizing its negative:
\begin{equation}
    \max_{y\in\mathcal{Y}} S_t(y)
    \;\equiv\;
    \min_{y\in\mathcal{Y}}
    \left\{
    \kappa y^2 - N_t y
    -\delta_t y^2
    -d y^3
    \right\}.
    \label{eq:add-central-bank-negative-score}
\end{equation}
Completing the square for the leading quadratic part gives
\begin{align}
    \kappa y^2 - N_t y
    &=
    \kappa
    \left(
    y-\frac{N_t}{2\kappa}
    \right)^2
    -
    \frac{N_t^2}{4\kappa}.
    \label{eq:add-central-bank-complete-square}
\end{align}
The last term depends only on the context and does not affect the minimizing
policy action. Dividing by the positive constant \(\kappa\) yields the displayed
latent-target form:
\begin{equation}
    \min_{y_t\in\mathcal{Y}}
    \left\{
    \left(
    y_t-\frac{N_t}{2\kappa}
    \right)^2
    +\varepsilon_t(y_t)
    \right\},
    \label{eq:add-central-bank-target-form}
\end{equation}
where
\begin{equation}
    \varepsilon_t(y_t)
    =
    -\frac{\delta_t}{\kappa}y_t^2
    -\frac{d}{\kappa}y_t^3
\end{equation}
collects residual decision-relevant deviations from the leading target-tracking
structure. Thus, the abstract law used in Figure~\ref{fig:central-bank}a is a
mechanism-level rewriting of the discovered symbolic score: policy follows a
latent target \(N_t/(2\kappa)\), with residual terms capturing state-dependent
curvature and higher-order deviations.

\subsubsection{Constrained \(U/C/T\) symbolic law}
\label{sec:add-central-bank-uct}

We further considered a constrained OptScientist Model whose optimization representation bottleneck contains three active components: an objective branch $U$, an inequality-constraint branch $C$, and a target-deviation or risk branch $T$. OptDiscover distills these components into symbolic form, yielding the final score
\begin{equation}
    S_t(y)=1.39737\,U_t(y)-1.39737\,C_t(y)-1.39737\,T_t(y).
    \label{eq:add-central-bank-uct-score}
\end{equation}
The three branches are
\begin{align}
U_t(y)
=
y\Big(&
0.02035\,a_t
+0.02035\,\pi^{\mathrm{core}}_t
+0.04071\,\Delta_t
-0.02035\,y
-0.51405
\Big),
\label{eq:add-central-bank-uct-u}
\end{align}
\begin{align}
C_t(y)
=
0.88835\,
\max\Big(0,\,
&0.4289\,a_t
-0.4289\,c_t
+0.4289\,g_t
+\Delta_t \notag\\
&-0.4289\,\ell_t
+0.4289\,v_t
-1.4289\,y
-0.92033
\Big),
\label{eq:add-central-bank-uct-c}
\end{align}
and
\begin{align}
T_t(y)
=
0.11839\Big(
&-0.55185\,a_t
+0.55185\,c_t
-0.55185\,\Delta_t
+\ell_t \notag\\
&-0.55185\,\tau_t
+0.55185\,v_t
+2y
\Big)^2 .
\label{eq:add-central-bank-uct-t}
\end{align}
Here \(\pi^{\mathrm{core}}_t\) is core PCE inflation and \(v_t\) is VIX. The
\(U\) branch captures policy pressure from financial conditions, core inflation
and policy momentum; the \(C\) branch captures a one-sided constraint involving
financial stress, growth and market volatility; and the \(T\) branch captures a
quadratic deviation penalty. The branch activations are identity for \(U\),
ReLU for \(C\) and square for \(T\). The final effective complexities are
\(19\), \(31\) and \(29\), respectively, giving total effective complexity
\(79\).

\subsubsection{Predictive performance and complexity}
\label{sec:add-central-bank-performance}

Table~\ref{tab:add-central-bank-performance} reports held-out test performance.
All uncertainty values are bootstrapped standard errors from \(10{,}000\)
quarter-level resamples with replacement. Parameter count follows the plotting
convention: for neural and tree models it counts model parameters or tree
capacity, whereas for symbolic models it counts distinct numerical constants in
the final expression. Symbolic structural complexity is reported separately.

\begin{table}[t]
\centering
\scriptsize
\setlength{\tabcolsep}{1.15pt}
\renewcommand{\arraystretch}{1.15}
\caption{
\textbf{Predictive performance on central-bank monetary-policy decisions.}
Exact accuracy is computed after mapping continuous predictions to the five
policy classes. Direction accuracy compares easing, no-change and tightening
directions. Values are mean \(\pm\) bootstrap standard error over 10,000
quarter-level resamples. OptScientist-C denotes OptScientist with constraints.
} \vspace{2mm}
\label{tab:add-central-bank-performance}
\begin{tabular}{@{}lcccccc@{}}
\toprule
Model & Params. & MSE$\downarrow$ & MAE$\downarrow$ & \(R^2\)$\uparrow$ & Exact acc.$\uparrow$ & Direction acc.$\uparrow$ \\
\midrule
MLP
& 5,057
& \(1.6093\pm0.6026\)
& \(0.9453\pm0.2250\)
& \(-0.1554\pm0.5005\)
& \(0.3571\pm0.1268\)
& \(0.4286\pm0.1313\) \\

Random forest
& 134,000
& \(1.0960\pm0.3639\)
& \(0.8340\pm0.1693\)
& \(0.2132\pm0.1663\)
& \(0.3571\pm0.1277\)
& \(0.5714\pm0.1323\) \\

XGBoost
& 6,600
& \(0.8501\pm0.2630\)
& \(0.7616\pm0.1387\)
& \(0.3897\pm0.1634\)
& \(0.3571\pm0.1284\)
& \(0.6429\pm0.1270\) \\

Neural EBM
& 5,121
& \(1.0276\pm0.4118\)
& \(0.6343\pm0.2132\)
& \(0.2622\pm0.3688\)
& \(0.5714\pm0.1327\)
& \(0.7143\pm0.1213\) \\

GBDT
& 6,600
& \(0.7595\pm0.2643\)
& \(0.6884\pm0.1430\)
& \(0.4547\pm0.1301\)
& \(0.5714\pm0.1314\)
& \(0.7857\pm0.1094\) \\

Symbolic regression
& 6
& \(0.6788\pm0.3428\)
& \(\mathbf{0.5636\pm0.1616}\)
& \(0.5126\pm0.2009\)
& \(0.5714\pm0.1322\)
& \(0.7143\pm0.1204\) \\

OptScientist
& 11
& \(\mathbf{0.6513\pm0.3080}\)
& \(0.5750\pm0.1521\)
& \(\mathbf{0.5324\pm0.1655}\)
& \(\mathbf{0.7857\pm0.1097}\)
& \(\mathbf{0.8571\pm0.0943}\) \\

OptScientist-C
& 11
& \(0.6792\pm0.3239\)
& \(0.5700\pm0.1605\)
& \(0.5124\pm0.2265\)
& \(0.7143\pm0.1217\)
& \(0.7857\pm0.1109\) \\
\bottomrule
\end{tabular}
\end{table}

OptScientist shows the strongest overall out-of-time predictive performance.
It attains the lowest MSE and highest \(R^2\), exact policy-stance accuracy and
direction accuracy, while symbolic regression achieves a slightly lower MAE.
OptScientist-C also remains competitive, with exact and direction accuracies of
\(0.7143\) and \(0.7857\), respectively.

Under decision-relevant structural-complexity accounting, the single-\(U\)
OptScientist law has effective complexity \(131\), the constrained \(U/C/T\)
law has total effective complexity \(79\), and the symbolic-regression baseline
has complexity \(21\). Thus, OptScientist-C yields a more compact and explicitly
structured law, whereas the single-\(U\) formulation achieves stronger
out-of-time performance on most metrics. This difference may arise because the
single-\(U\) formulation retains richer state-dependent interactions and
higher-order decision structure, whereas the constrained \(U/C/T\) decomposition
imposes stronger structural restrictions. These restrictions improve parsimony
and interpretability, but may omit variation that remains predictive in unseen
periods.

\subsection{Token-Efficient Model Discovery}
\label{sec:add-token-efficiency}

We provide additional details for the token-efficient model-discovery experiment in Sec.~\ref{sec:token_efficiency}. OptScientist is compared with three LLM-based model-discovery baselines: (i) LLM-only discovery, which directly proposes and iteratively revises candidate models; (ii) GeCCo, a code-assisted LLM approach \citep{rmus2026generating}; and (iii) MDA, a Bayesian experiment-design approach \citep{murphy2026model}. All methods use the same DeepSeek-V4-Flash \citep{deepseekai2026deepseekv4} as the underlying language model and receive the same research description, variable information, and training data. The comparison is conducted on the CPC18 dataset and the central-bank monetary-policy dataset described in Secs.~\ref{sec:expr-human-choice} and~\ref{sec:expr-central-bank}. For each LLM-based baseline, we evaluate maximum generation settings of \(8{,}192\), \(12{,}000\), \(16{,}384\) and \(24{,}000\) tokens per LLM call.

\textbf{Evaluation protocol.}
Model generation, revision and selection use only the training and validation data. The held-out test set is not exposed during the discovery process and is evaluated after the final model has been selected. We record the held-out predictive accuracy of each discovered model together with cumulative LLM output-token consumption across the complete discovery run. The maximum-token setting denotes the generation limit for each individual LLM call and therefore differs from the cumulative number of generated tokens reported below.

\textbf{Token accounting.}
We use cumulative output-token consumption as the primary measure of language-model budget. The discovery methods differ in their input representations, prompting structures and intermediate context, making input-token counts less directly comparable across approaches. Generated output tokens provide a common measure of the model-generation budget and are also the more expensive component under the API pricing used in these experiments. We additionally report held-out accuracy per output token as a descriptive summary of the accuracy--token trade-off. The analysis concerns LLM-token efficiency; training and symbolic-search computation outside the language model are not included in this measure.

\begin{table}[t]
\centering
\scriptsize
\setlength{\tabcolsep}{6pt}
\renewcommand{\arraystretch}{1.18}
\caption{
\textbf{Token-efficient model discovery on the CPC18 dataset.}
Held-out predictive accuracy and cumulative LLM output-token consumption are reported for each discovery setting. The maximum-token column gives the generation limit per LLM call. OptScientist achieves held-out accuracy near the apparent CPC18 predictive
ceiling with substantially lower output-token consumption.
} \vspace{2mm}
\label{tab:add-token-cpc18}
\begin{tabular}{@{}lrrrr@{}}
\toprule
Model & Max tokens & Output tokens & Held-out acc.$\uparrow$ & Acc./token \((\times 10^{-5})\)$\uparrow$ \\
\midrule
LLM-only   & 8,192  & 17,981 & 83.354\% & 4.636 \\
   & 12,000 & 15,771 & 82.956\% & 5.260 \\
   & 16,384 & 16,048 & 83.173\% & 5.183 \\
   & 24,000 & 16,124 & 83.259\% & 5.164 \\
GeCCo \citep{rmus2026generating}     & 8,192  & 35,947 & 83.777\% & 2.331 \\
      & 12,000 & 37,317 & 83.487\% & 2.237 \\
      & 16,384 & 35,873 & 83.667\% & 2.332 \\
      & 24,000 & 35,760 & 83.540\% & 2.336 \\
MDA  \citep{murphy2026model}      & 8,192  & 45,761 & 83.114\% & 1.816 \\
        & 12,000 & 44,807 & 83.346\% & 1.860 \\
        & 16,384 & 41,550 & 83.395\% & 2.007 \\
        & 24,000 & 44,516 & 83.172\% & 1.868 \\
\midrule
\textbf{OptScientist} & -- & \textbf{6,396} & 82.900\% & \textbf{12.961} \\
\bottomrule
\end{tabular}
\end{table}

\begin{table}[htbp]
\centering
\scriptsize
\setlength{\tabcolsep}{6pt}
\renewcommand{\arraystretch}{1.18}
\caption{
\textbf{Token-efficient model discovery on central-bank monetary-policy decisions.}
Held-out policy-stance accuracy and cumulative LLM output-token consumption are reported for each discovery setting. OptScientist achieves the highest held-out point estimate while using substantially fewer output tokens than the LLM-based discovery baselines.
}\vspace{2mm}
\label{tab:add-token-central-bank}
\begin{tabular}{@{}lrrrr@{}}
\toprule
Model & Max tokens & Output tokens & Held-out acc.$\uparrow$ & Acc./token \((\times 10^{-5})\)$\uparrow$ \\
\midrule
LLM-only   & 8,192  & 18,381 & 44.318\% & 2.411 \\
   & 12,000 & 16,691 & 49.432\% & 2.962 \\
   & 16,384 & 17,361 & 47.727\% & 2.749 \\
   & 24,000 & 18,142 & 40.341\% & 2.224 \\
GeCCo \citep{rmus2026generating}     & 8,192  & 23,188 & 64.205\% & 2.769 \\
      & 12,000 & 26,350 & 56.818\% & 2.156 \\
      & 16,384 & 25,447 & 56.250\% & 2.210 \\
      & 24,000 & 26,677 & 53.409\% & 2.002 \\
MDA \citep{murphy2026model}       & 8,192  & 29,611 & 68.182\% & 2.303 \\
        & 12,000 & 27,687 & 64.205\% & 2.319 \\
        & 16,384 & 25,576 & 70.455\% & 2.755 \\
        & 24,000 & 29,373 & 64.773\% & 2.205 \\
\midrule
\textbf{OptScientist} & -- & \textbf{5,976} & \textbf{78.600\%} & \textbf{13.153} \\
\bottomrule
\end{tabular}
\end{table}

\textbf{Token-efficiency results.}
On human decisions under risk and ambiguity, the LLM-based baselines achieve closely clustered held-out accuracies while consuming between \(15{,}771\) and \(45{,}761\) output tokens. Notably, the strongest predictive models cluster around 82--84\% held-out
accuracy. Performance on the CPC18 dataset appears to saturate across all methods at approximately 83\% held-out accuracy, suggesting that CPC18 poses a relatively less challenging model-discovery problem. Nevertheless, OptScientist reaches competitive held-out accuracy using \(6{,}396\) output tokens, corresponding to \(59.4\%\) fewer tokens than the lowest-token baseline. Its accuracy per output token is \(12.961\times10^{-5}\), compared with a maximum of \(5.260\times10^{-5}\) among the LLM-based baselines.

On central-bank monetary-policy decisions, OptScientist achieves the highest held-out policy-stance accuracy, reaching \(78.6\%\) with \(5{,}976\) output tokens. The strongest LLM-based baseline reaches \(70.455\%\) accuracy with \(25{,}576\) output tokens, while output-token consumption across the baseline settings ranges from \(16{,}691\) to \(29{,}611\). Compared with the CPC18 dataset, modeling monetary-policy decisions appears more challenging for the LLM-based baselines, as the discovered models exhibit greater variability in performance. OptScientist also reaches an accuracy-per-token value of \(13.153\times10^{-5}\), compared with a maximum of \(2.962\times10^{-5}\) among the baselines. Together, the two tasks show that OptScientist reaches competitive or stronger held-out predictive performance with substantially less LLM generation, supporting the accuracy--token-efficiency trade-off reported in Figure~\ref{fig:llm_baseline}.

\end{document}